\documentclass[10pt,twocolumn,letterpaper]{article}

\usepackage[pagenumbers]{cvpr} 

\definecolor{cvprblue}{rgb}{0.21,0.49,0.74}
\usepackage[pagebackref,breaklinks,colorlinks,allcolors=cvprblue]{hyperref}

\PassOptionsToPackage{table}{xcolor}

\usepackage{graphicx}
\usepackage{booktabs}
\usepackage{array}
\usepackage{tabularx}
\usepackage{multirow}
\usepackage{makecell}
\usepackage{colortbl}
\usepackage{adjustbox}
\usepackage{threeparttable}
\usepackage{longtable}
\usepackage{subcaption}
\usepackage{rotating}
\usepackage{wrapfig}
\usepackage{cuted}

\usepackage{amsmath}
\usepackage{amsfonts}
\usepackage{nicefrac}
\usepackage{microtype}

\usepackage{newtxmath}

\usepackage[T1]{fontenc}
\usepackage[utf8]{inputenc}

\usepackage{url}
\usepackage{enumitem}
\usepackage{ragged2e}
\usepackage{quoting}
\usepackage{framed}
\usepackage{listings}

\usepackage{pifont}
\usepackage{fontawesome5}
\usepackage{academicons}

\usepackage{CJKutf8}

\usepackage{xcolor}

\definecolor{darkgreen}{RGB}{0,100,0}
\definecolor{darkred}{RGB}{139,0,0}

\definecolor{gold}{RGB}{212,175,55}

\definecolor{tabblue}{HTML}{1F77B4}
\definecolor{taborange}{HTML}{FF7F0E}
\definecolor{tabgreen}{HTML}{2CA02C}
\definecolor{tabred}{HTML}{D62728}
\definecolor{tabpurple}{HTML}{9467BD}
\definecolor{tabpink}{HTML}{FF0080}

\definecolor{SATTbg}{RGB}{250,244,244}

\newcolumntype{C}{>{\centering\arraybackslash}X}

\newcommand{\cmark}{%
    \textcolor{green!45!black}{\ding{51}}%
}

\newcommand{\xmark}{%
    \textcolor{gray!55}{\ding{55}}%
}

\definecolor{lapsgreen}{RGB}{0,85,55}
\definecolor{lapsred}{RGB}{125,35,35}

\definecolor{tabpink}{HTML}{ff0080}

\usepackage[capitalize,noabbrev]{cleveref}

\providecolor{gold}{RGB}{212,175,55}
\providecolor{darkred}{RGB}{150,30,30}
\providecolor{lapsgreen}{RGB}{0,85,55}
\providecommand{\cmark}{\ensuremath{\checkmark}}
\providecommand{\xmark}{\ensuremath{\times}}

\providecolor{oursblue}{RGB}{40,90,134}
\providecolor{oursbg}{RGB}{232,240,248}

\providecommand{\ourscell}[1]{%
    \cellcolor{oursbg}#1%
}

\providecommand{\cmark}{\ensuremath{\checkmark}}
\providecommand{\xmark}{\ensuremath{\times}}

\usepackage{lipsum}

\def\confName{CVPR}
\def\confYear{2027}

\title{\textit{PanoUAV-Track}: Full-Spherical UAV Visual Tracking in the Wild}

\author{
\begin{tabular}{c}
Bingyu Li$^{1}$ \thanks{Work done during an internship at TeleAI.} \quad Lichang Yang$^{2}$  \quad Zhiyuan Zhao$^{3}$ \quad Junyu Gao$^{3, 2}$ \quad Xuelong Li$^{3}$ \thanks{Corresponding Author}\\
$^{1}$University of Science and Technology of China\quad
$^{2}$Northwestern Polytechnical University \\
$^{3}$The Institute of Artificial Intelligence (TeleAI) of China Telecom
\end{tabular}
}

\begin{document}
\maketitle

\begin{abstract}
Existing UAV tracking benchmarks predominantly rely on
perspective cameras with limited fields of view.
Omnidirectional cameras enable full-spherical observation,
yet tracking from moving UAV platforms remains largely
unexplored.
Unlike ground-level panoramic tracking,
UAV-mounted omnidirectional cameras observe targets
across both azimuth and elevation, introducing
geometric challenges under equirectangular projection (ERP).
To investigate this setting, we capture approximately
\textbf{40 hours of real-world 8K panoramic UAV video}
and construct \textbf{PanoUAV-Track} through
temporal clip extraction, target annotation,
and the integration of simulated sequences.
The resulting benchmark comprises
1,895 sequences, 507,961 frames,
and 2,872 annotated objects,
covering real-world daytime and nighttime scenes
alongside simulation, and supporting both visual
and language-based target specification.
Through extensive evaluations, we reveal a clear
\textbf{perspective-to-spherical} generalization gap
and identify tracking degradation associated with
latitude-dependent distortion and ERP seam discontinuity.
To address these challenges, we propose
\textbf{PanoUAV-Tracker (PUTrack)}, a geometry-aware
tracking framework combining latitude-adaptive search
and seam-aware periodic sampling.
PUTrack improves zero-shot and in-domain tracking,
particularly for seam-crossing targets.
Code and benchmark resources will be available\footnote{\textcolor{tabpink}{https://libingyu01.github.io/PanoUAV-Track/}}.
\end{abstract}

\section{Introduction}
\label{sec:introduction}

\begin{figure}[t]
    \centering
    \includegraphics[width=\linewidth]{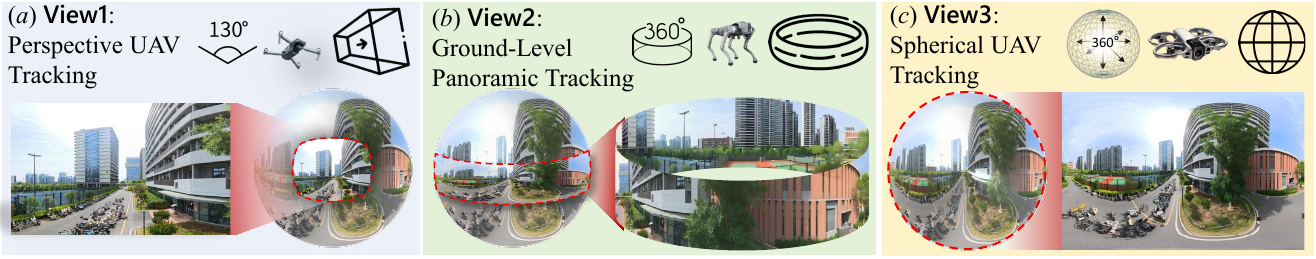}
    \includegraphics[width=\linewidth]{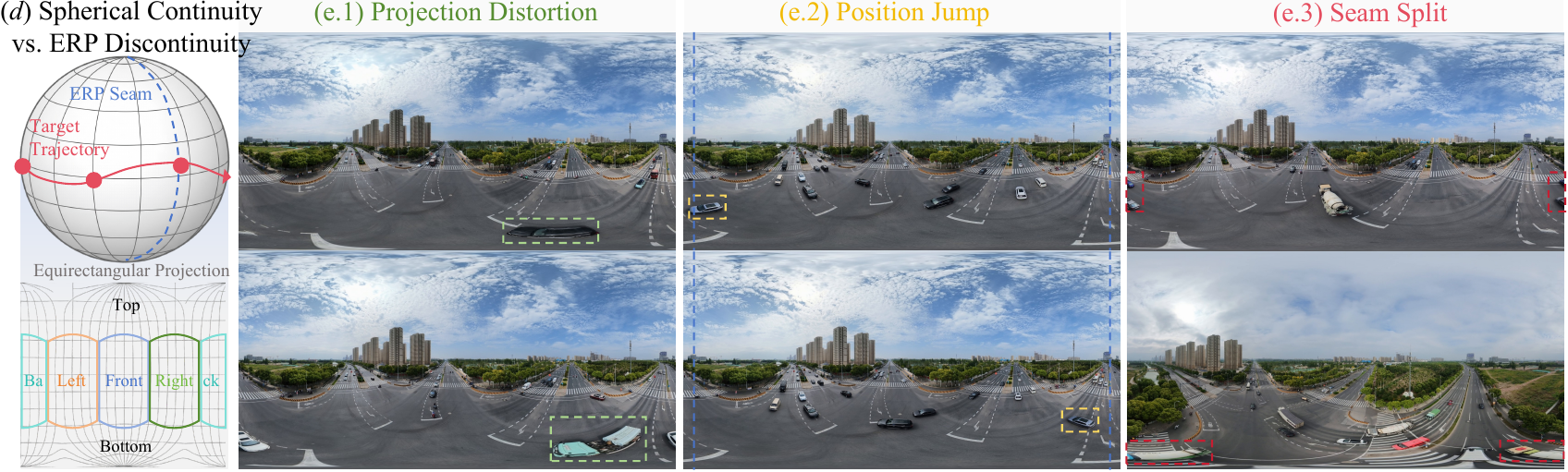}
    \caption{
    \textbf{Observation settings and projection challenges.}
    (a--c) Perspective UAV, ground-level panoramic, and
    full-spherical UAV tracking.
    (d) Spherical continuity versus ERP discontinuity.
    (e.1--e.3) Projection distortion, position jumps, and seam splitting.
    }
    \label{fig:motivation}
    \vspace{-4.5mm}
\end{figure}

Visual object tracking from unmanned aerial vehicles
(UAVs) has witnessed substantial progress in recent
years, driven by advances in tracking algorithms
and aerial tracking benchmarks
\citep{mueller2016benchmark,zhang2022webuav,li2024uavnlt,
wu2025ort,xue2025sglatrack,AIFlowAn,AIFlowShao}.
However, existing benchmarks and tracking methods
predominantly rely on perspective cameras with
limited fields of view\cite{li2024uavnlt,li2025stitchfusion,li2026exploring,zhang2022webuav}.
Under such observations, targets can easily disappear
from the image due to UAV motion or changes in viewing
direction, even when they remain within the surrounding
environment (\cref{fig:motivation}(a)).
This limitation raises a fundamental question:
\textit{can visual trackers developed for perspective
observations effectively generalize to omnidirectional
UAV tracking?}

Omnidirectional cameras offer a promising solution
by capturing the surrounding environment across
a complete spherical field of view.
Existing panoramic benchmarks, such as
360VOT~\citep{huang2023_360vot},
360VOTS~\citep{xu2025_360vots}, and
PanoVOS~\citep{yan2024panovos},
have advanced object tracking and segmentation
under omnidirectional observations.
However, these benchmarks primarily focus on
general panoramic scenarios rather than
dynamic observations from aerial platforms
(\cref{fig:motivation}(b)).
Unlike ground-level panoramic tracking,
UAV-mounted omnidirectional cameras experience
continuously changing aerial viewpoints,
where targets may appear at arbitrary azimuths
and elevations (\cref{fig:motivation}(c)).
We refer to this challenging setting as
\textbf{full-spherical UAV object tracking}.

Beyond extending the field of view,
full-spherical observation introduces
fundamentally different tracking challenges.
Omnidirectional videos are typically represented
using equirectangular projection (ERP),
which maps spherical observations onto
a rectangular image plane.
However, this representation does not preserve
the geometric properties of the sphere.
As illustrated in \cref{fig:motivation}(d, e.1--e.3),
target appearance varies with latitude under ERP,
while seam-crossing targets may appear on
opposite sides of the image.
Consequently, a target that moves continuously
on the sphere may exhibit substantial
appearance deformation or abrupt displacement
in the projected image.
These effects challenge conventional tracking
assumptions and raise concerns about the
generalization of perspective-based trackers
to full-spherical observations.

To systematically investigate these challenges,
we introduce \textbf{PanoUAV-Track},
a large-scale benchmark for full-spherical
UAV visual tracking.
The benchmark comprises
\textbf{1,895 sequences, 507,961 frames,
and 2,872 annotated objects},
covering 951 daytime, 606 nighttime,
and 338 simulated sequences.
It provides diverse aerial viewpoints,
target trajectories, and observation conditions,
together with visual and language-based
target specifications.
Based on this benchmark, we investigate
three fundamental research questions:
\textbf{\textcolor{blue}{(RQ1)}}
How well do perspective trackers generalize
to full-spherical UAV observations?
\textbf{\textcolor{blue}{(RQ2)}}
Does the generalization gap persist across
different observation domains?
\textbf{\textcolor{blue}{(RQ3)}}
How does spherical projection geometry
affect tracking performance?
Our comprehensive evaluations reveal
a substantial perspective-to-spherical
generalization gap, with tracking performance
particularly affected by latitude-dependent
distortion and ERP seam discontinuities.

Motivated by these findings, we further propose
\textbf{PanoUAV-Tracker (PUTrack)},
a geometry-aware tracking framework
specifically designed for full-spherical
UAV observations.
PUTrack combines template-guided target
localization with geometry-aware search
construction.
Its Geometry-Aware Search Adapter
(\textbf{GASA}) incorporates
Latitude-Adaptive Patch Sampling (\textbf{LAPS})
to compensate for latitude-dependent distortion
and Seam-Aware Search Cropping (\textbf{SASC})
to preserve spatial continuity across
panoramic boundaries.
Thus, PUTrack maintains coherent target localization
in ERP coordinates.

Our main contributions are summarized as follows:

\begin{itemize}

\item \textbf{Full-spherical UAV tracking.}
We investigate visual tracking using UAV-mounted
omnidirectional cameras, extending aerial tracking
to full-spherical observations.

\item \textbf{PanoUAV-Track benchmark and analysis.}
PanoUAV-Track supports visual and language-based tracking.
Extensive evaluations reveal the limitations
of perspective trackers under full-spherical observations.

\item \textbf{PanoUAV-Tracker (PUTrack).}
A geometry-aware tracking framework is developed
with latitude-adaptive search and periodic seam-aware
sampling.

\end{itemize}
\section{Related Work}
\label{sec:related_work}

\subsection{Perspective Tracking}
\indent

Visual object tracking has progressed from Siamese architectures to Transformer- and sequence-based frameworks, supported by large-scale benchmarks such as TrackingNet~\citep{muller2018trackingnet}, GOT-10k~\citep{huang2019got}, and LaSOT~\citep{fan2019lasot}.
Recent methods further strengthen target modeling and temporal reasoning
\citep{chen2021transformer,yan2021learning,cui2022mixformer,ye2022joint,
chen2023seqtrack,cai2024hiptrack,liang2025autoregressive}.
UAV benchmarks extend this setting to aerial observation
\citep{mueller2016benchmark,zhang2022webuav,li2024uavnlt},
while vision--language tracking introduces natural-language target specification
\citep{wang2021tnl2k,feng2021snlt,zhou2023jointnlt,ma2023decoupletnl,
ma2024uvltrack,shao2024querynlt,liu2025mambavlt,li2025dutrack}.
Despite differences in task formulation and model design, these benchmarks remain predominantly based on perspective imagery.
PanoUAV-Track complements them by evaluating both visual and language target specification under full-spherical UAV observation.

\begin{figure}[t]
    \centering
    \includegraphics[width=\linewidth]{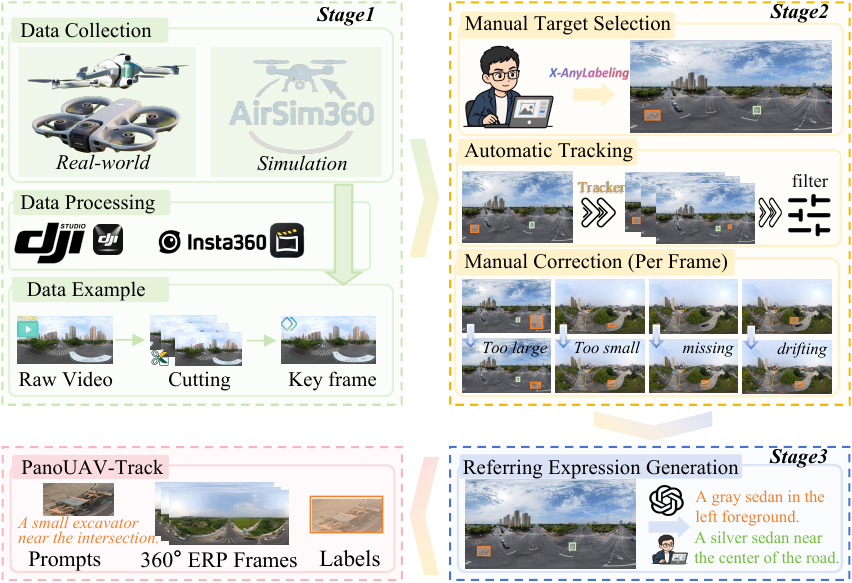}
    \caption{
    \textbf{Construction pipeline of PanoUAV-Track.}
    Tracking clips are constructed through manual initialization, automatic
    propagation, frame-level correction, and language annotation.
    }
    \label{fig:annotation_pipeline}
\end{figure}


\begin{table*}[t]
\centering

\caption{
\textbf{Comparison of PanoUAV-Track with representative visual tracking benchmarks.}
}
\label{tab:dataset_comparison}

\vspace{-2mm}

\scriptsize
\setlength{\tabcolsep}{6pt}
\renewcommand{\arraystretch}{0.99}
\setlength{\arrayrulewidth}{0.4pt}

\resizebox{\textwidth}{!}{%
\begin{tabular}{c|lcccccccl}

\noalign{\global\setlength{\arrayrulewidth}{0.6pt}}
\hline
\noalign{\global\setlength{\arrayrulewidth}{0.4pt}}

\rowcolor{gray!12}
Type &
Dataset &
Platform &
Observation &
Visual &
Language &
Sequences &
Frames &
Data &
Publication \\

\hline


\multirow[c]{4}{*}{%
\shortstack[c]{General-purpose\\Tracking}}

& TrackingNet~\citep{muller2018trackingnet}
& General & Perspective
& \cmark & \xmark
& 30K+ & 14M+
& Real & ECCV'18 \\

& GOT-10k~\citep{huang2019got}
& General & Perspective
& \cmark & \xmark
& 10K+ & 1.5M+
& Real & TPAMI'21 \\

& LaSOT~\citep{fan2019lasot}
& General & Perspective
& \cmark & \cmark
& 1,400 & 3.5M+
& Real & CVPR'19 \\

& TNL2K~\citep{wang2021tnl2k}
& General & Perspective
& \cmark & \cmark
& 2,000 & 1.24M
& Real+Sim. & CVPR'21 \\

\hline


\multirow[c]{2}{*}{%
\shortstack[c]{UAV\\Tracking}}

& UAV123~\citep{mueller2016benchmark}
& UAV & Perspective
& \cmark & \xmark
& 123 & 112.6K
& Real+Sim. & ECCV'16 \\

& WebUAV-3M~\citep{zhang2022webuav}
& UAV & Perspective
& \cmark & \cmark
& 4,500 & 3.3M+
& Real & TPAMI'23 \\

\hline


\multirow[c]{5}{*}{%
\shortstack[c]{Panoramic\\Tracking}}

& 360VOT~\citep{huang2023_360vot}
& Panorama & 360$^\circ$ ERP
& \cmark & \xmark
& 120 & 113K
& Real & ICCV'23 \\

& 360VOS (360VOTS)~\citep{xu2025_360vots}
& Panorama & 360$^\circ$ ERP
& \cmark & \xmark
& 290 & 242K
& Real+Sim. & TPAMI'25 \\

& PanoVOS~\citep{yan2024panovos}
& Panorama & 360$^\circ$ ERP
& \cmark & \xmark
& 150 & 14.0K
& Real & ECCV'24 \\

& QuadTrack~\citep{luo2025omnidirectional}
& Quadruped & 360$^\circ$ ERP
& \cmark & \xmark
& 32 & 19.2K
& Real & CVPR'25 \\


& \ourscell{\textbf{PanoUAV-Track}}
& \ourscell{UAV}
& \ourscell{360$^\circ$ $\times$ 180$^\circ$ ERP}
& \ourscell{\cmark}
& \ourscell{\cmark}
& \ourscell{\textbf{1,895}}
& \ourscell{\textbf{507.96K}}
& \ourscell{Real + Sim.}
& \ourscell{Ours} \\

\noalign{\global\setlength{\arrayrulewidth}{0.6pt}}
\hline
\noalign{\global\setlength{\arrayrulewidth}{0.4pt}}

\end{tabular}%
}

\vspace{-3mm}
\end{table*}

\begin{figure*}[t]
    \centering
    \includegraphics[width=\linewidth]{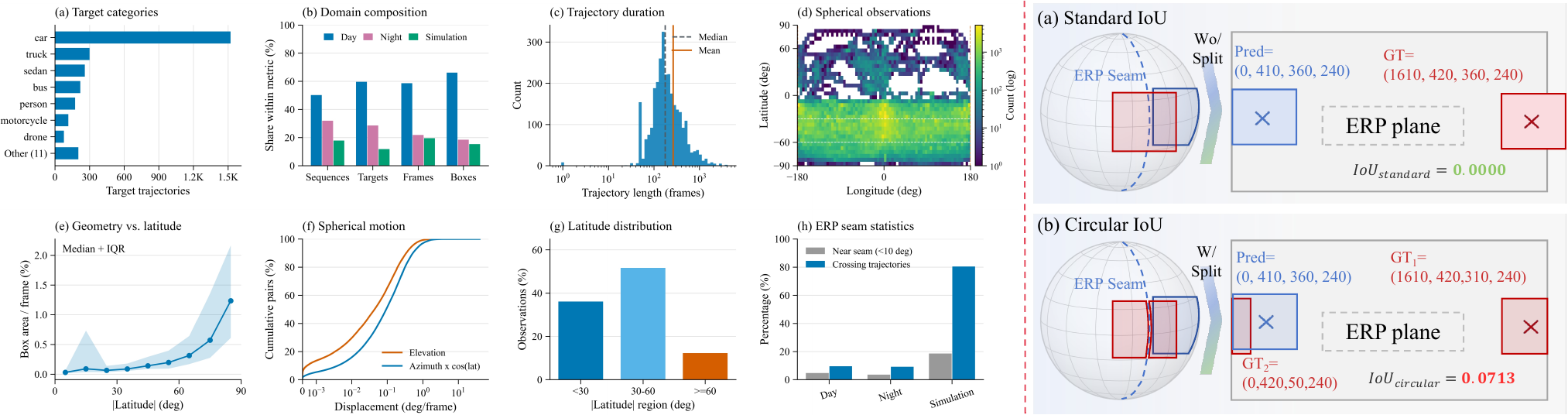}
    \vspace{-2em}
    \caption{
    Overview of PanoUAV-Track dataset statistics and circular IoU.
    \textbf{Left:} Dataset statistics covering target diversity,
    trajectory characteristics, and spherical distributions.
    \textbf{Right:} Illustration of circular IoU, which handles
    seam-crossing bounding boxes.
    }
    \label{fig:xx1}
    \vspace{-1em}
\end{figure*}

\subsection{Omnidirectional and Spherical Tracking}
\indent

Omnidirectional tracking extends visual observation
beyond perspective cameras while introducing
spherical projection challenges.
360VOT~\citep{huang2023_360vot} establishes an
omnidirectional single-object tracking benchmark
with sphere-aware representations and evaluation.
360VOTS~\citep{xu2025_360vots} extends this setting
to tracking and segmentation, while
PanoVOS~\citep{yan2024panovos} explores temporal
consistency in panoramic video segmentation.
Recent studies further investigate omnidirectional
tracking on mobile robotic platforms.
JRDB-PanoTrack~\citep{le2024jrdb} provides
panoptic tracking annotations for robot-centered
environments, while
OmniTrack~\citep{luo2025omnidirectional}
introduces QuadTrack for omnidirectional
multi-object tracking on quadruped robots.
However, these studies primarily focus on
general panoramic videos or ground-level robotics.
In contrast, PanoUAV-Track investigates
full-spherical tracking from moving UAVs,
characterized by varying aerial viewpoints
and target motion across the spherical domain.

\section{Full-Spherical UAV Tracking}
\label{sec:problem}

\subsection{Problem Formulation}
\indent

To maintain compatibility with conventional visual object tracking,
we formulate omnidirectional UAV tracking in the equirectangular
projection (ERP) image.
Let $\mathcal{V}=\{I_t\}_{t=1}^{T}$ denote an omnidirectional UAV
video sequence consisting of $T$ ERP frames.
Unlike conventional perspective tracking, which operates on a locally
continuous image plane, ERP maps the spherical viewing domain
$\mathbb{S}^{2}$ onto a rectangular image, introducing horizontal
periodicity and projection distortions.
For a pixel location $(u,v)$ in an ERP frame of width $W$ and
height $H$, the corresponding spherical longitude and latitude
are defined as $\lambda = \frac{2\pi u}{W}-\pi,\phi = \frac{\pi}{2}-\frac{\pi v}{H}$.

At each time step $t$, the target is represented by an axis-aligned
bounding box $B_t=(x_t,y_t,w_t,h_t)$, where $(x_t,y_t)$ denotes
the top-left corner and $(w_t,h_t)$ specifies its width and height.
The bounding-box center is further projected onto the sphere
to obtain its corresponding spherical coordinates
$(\lambda_t,\phi_t)$.


\subsection{The PanoUAV-Track Benchmark}
\label{sec:benchmark}
\indent

\textbf{Data Collection and Processing.}
As illustrated in \cref{fig:annotation_pipeline},
the construction of PanoUAV-Track follows a three-stage
pipeline. We collect approximately 40 hours of real-world
8K omnidirectional videos using UAV-mounted panoramic
cameras, supplemented with simulated aerial sequences.
All videos are processed and converted
into $360^\circ\times180^\circ$ equirectangular projection
(ERP) format. The resulting recordings are further
segmented into tracking clips based on scene continuity
and target visibility.

\textbf{Trajectory Annotation and Quality Control.}
We adopt a semi-automatic annotation pipeline combining
manual initialization, automatic tracking, and human
verification. For each sequence, annotators first select
the target and manually specify its initial bounding box.
An existing tracker then propagates the target state
across subsequent frames, followed by a filtering step
to refine the initial trajectory. To ensure annotation
quality, all generated trajectories undergo frame-level
manual inspection and correction, particularly addressing
inaccurate bounding-box sizes and missing annotations. Additional attention is paid to
projection distortions and targets crossing the ERP seam.
Automatic tracking serves only as an annotation aid,
while the final ground-truth trajectories are determined
through human verification and correction.

\textbf{Referring Expression Annotation.}
To support language-guided tracking, we further associate
each target trajectory with a natural-language referring
expression. Candidate descriptions are first generated
with the assistance of Qwen3 \cite{yang2025qwen3} based on
the target and its surrounding visual context.
These descriptions are subsequently reviewed and
manually corrected at the sequence level to ensure
target specificity and reduce ambiguity.

The resulting benchmark provides aligned ERP video frames,
frame-level bounding-box annotations, and language prompts.
The same ground-truth trajectories are shared
across visual-prompt, language-prompt, and joint
visual--language tracking protocols, enabling consistent
evaluation under different target-specification settings.

\begin{figure*}[t]
    \centering
    \includegraphics[width=0.9\linewidth]{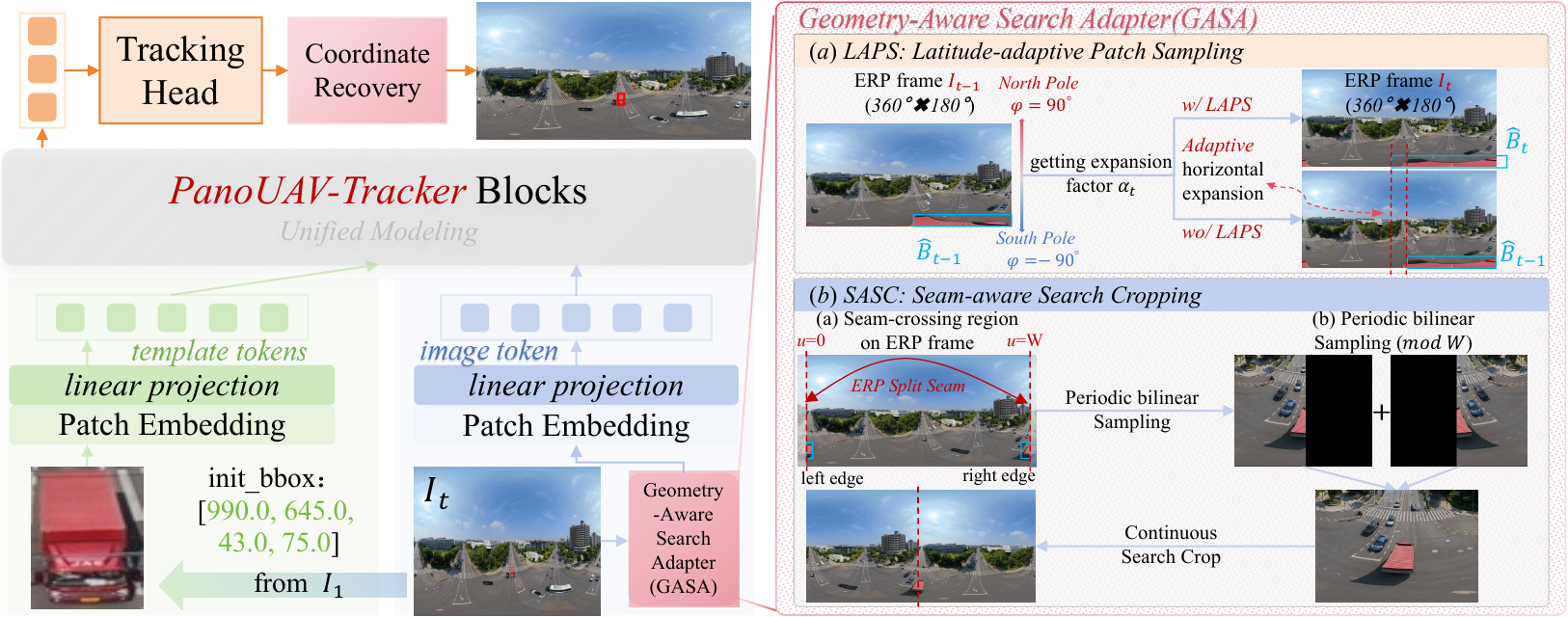}
    \vspace{-1em}
    \caption{
    \textbf{Overview of the proposed PanoUAV-Tracker.}
    \textbf{Left:} The overall framework enables geometry-aware
    visual tracking in omnidirectional UAV imagery.
    \textbf{Right:} LAPS compensates for latitude-dependent ERP
    distortion, while SASC preserves spatial continuity.
    }
    \label{fig:gasa_method}
    \vspace{-1em}
\end{figure*}

\subsection{Dataset Overview and Comparison}
\indent

Table~\ref{tab:dataset_comparison} compares PanoUAV-Track
with representative visual tracking benchmarks.
Existing UAV tracking datasets primarily rely on
perspective observations, while panoramic tracking
benchmarks mainly focus on general omnidirectional
scenarios. In contrast, PanoUAV-Track combines
UAV-mounted full-spherical observations with both
visual and language-based target specifications.
The benchmark contains 1,895 sequences, 507,961 frames,
and 2,872 annotated objects across real-world daytime,
nighttime, and simulated environments, exceeding the
existing panoramic benchmarks in the table in both
sequence and frame counts.

\cref{fig:xx1} presents the statistical
characteristics of PanoUAV-Track.
The dataset covers diverse target categories and
trajectory durations across different domains.
The spherical observation and motion distributions
demonstrate diverse viewing directions and target
displacements, while the latitude-dependent variation
in bounding-box area reflects the geometric distortions
introduced by ERP projection.
Furthermore, the presence of seam-crossing trajectories,
particularly in simulated environments, highlights the
challenges of horizontal periodicity in omnidirectional
tracking. These characteristics motivate seam-aware
evaluation, as illustrated by the circular IoU comparison
in \cref{fig:xx1}.


\subsection{Evaluation Protocol}
\label{sec:evaluation}
\indent

We adopt \emph{Success AUC} and \emph{Normalized Precision}
($P_{\mathrm{Norm}}$) for image-space tracking evaluation,
following the standard evaluation procedures used in
classic visual object tracking.
Specifically, Success AUC is computed using
\textbf{circular IoU}, as illustrated in \cref{fig:xx1} (right),
which accounts for the horizontal periodicity of ERP images
and properly handles bounding boxes crossing the projection seam.

To further evaluate tracking performance in the spherical
domain, we additionally report \textbf{Angular Precision}
($P_{\mathrm{Ang}}$).
The predicted and ground-truth target centers are mapped
from ERP coordinates onto the unit sphere.
Given spherical coordinates $(\lambda,\phi)$, the
corresponding unit vector is defined as
\begin{equation}
\mathbf{s}(\lambda,\phi) =
\begin{bmatrix}
\cos\phi\cos\lambda \\
\cos\phi\sin\lambda \\
\sin\phi
\end{bmatrix}.
\end{equation}
The angular localization error at frame $t$ is
computed as
\begin{equation}
\theta_t =
\arccos\left(
\operatorname{clip}
\left(
\mathbf{s}_t^p \cdot \mathbf{s}_t^g,
-1,1
\right)
\right),
\end{equation}
where $\mathbf{s}_t^p$ and $\mathbf{s}_t^g$ denote
the predicted and ground-truth unit vectors,
respectively.
Angular Precision is then defined as
\begin{equation}
P_{\mathrm{Ang}} =
\frac{1}{N}
\sum_{t=1}^{N}
\mathbb{I}\left[
\theta_t \leq \tau_{\mathrm{Ang}}
\right],
\label{eq:angular_precision}
\end{equation}
where $N$ is the number of evaluated frames,
$\mathbb{I}[\cdot]$ is the indicator function,
and $\tau_{\mathrm{Ang}}=3^\circ$ is the angular
error threshold.
This metric remains continuous across the ERP seam
and directly measures spherical localization accuracy.


\section{PanoUAV-Tracker}
\label{sec:method}
\indent

We propose \textbf{PanoUAV-Tracker (PUTrack)},
a geometry-aware visual tracking framework
for full-spherical UAV observations.
Unlike conventional perspective tracking,
ERP introduces latitude-dependent geometric
distortion and horizontal boundary discontinuities,
leading to inconsistent search coverage,
localization drift, and fragmented target
observations, as shown in \cref{fig:motivation}.
PUTrack addresses these challenges through
a unified tracking pipeline that combines
template-guided target localization,
geometry-aware search construction,
and consistent coordinate recovery.
The overall framework is illustrated in
\cref{fig:gasa_method}.


\subsection{Overall Framework}
\label{sec:framework}
\indent

Given an ERP video sequence
$\mathcal{V}=\{I_t\}_{t=1}^{T}$
and an initial target bounding box $B_1$,
PUTrack aims to estimate the target
bounding box $\widehat{B}_t$
at each subsequent frame.
PUTrack adopts the one-stream template--search
tracking paradigm of OSTrack~\citep{ye2022joint},
while explicitly modeling panoramic geometry
during search construction and target localization.

The initial target observation in $I_1$
provides a fixed appearance template $Z_1$.
For each subsequent frame $t\geq2$,
PUTrack determines the search region from
the previous predicted bounding box
$\widehat{B}_{t-1}$,
with $\widehat{B}_1=B_1$.
The search center $(c_x,c_y)$ and
the original search-region side length $L_t$
are derived from $\widehat{B}_{t-1}$.
PUTrack then constructs a geometry-aware
search image $S_t$ through
Latitude-Adaptive Patch Sampling (\cref{sec:laps})
and Seam-Aware Search Cropping (\cref{sec:sasc}).

The template $Z_1$ and search image $S_t$
are jointly processed by the tracking network:

\begin{equation}
\widehat{B}_t^{s}
=
\left(
\widehat{p}_t,
\widehat{q}_t,
\widehat{w}_t^{s},
\widehat{h}_t^{s}
\right)
=
\mathcal{F}(Z_1,S_t),
\label{eq:putrack_forward}
\end{equation}

where $\mathcal{F}$ denotes the
template--search tracking network,
and $\widehat{B}_t^{s}$ represents
the predicted bounding box in
normalized search coordinates.

The prediction is subsequently transformed
back into ERP coordinates through
geometry-consistent coordinate recovery.
The recovered bounding box $\widehat{B}_t$
determines the search region for frame $t+1$,
forming an autoregressive tracking process.


\subsection{Latitude-Adaptive Patch Sampling}
\label{sec:laps}
\indent

As illustrated in \cref{fig:gasa_method}(a),
ERP introduces nonuniform geometric
distortion across different latitudes.
Consequently, conventional fixed-scale
search regions provide inconsistent
spherical coverage as target latitude varies.
Given the previous prediction
$\widehat{B}_{t-1}
=(\widehat{x}_{t-1},
\widehat{y}_{t-1},
\widehat{w}_{t-1},
\widehat{h}_{t-1})$,
PUTrack obtains its center
$(c_x,c_y)$ and the original
search-region side length $L_t$.
Following the ERP mapping defined
in Sec.~\ref{sec:problem},
the predicted target latitude is

\begin{equation}
\widehat{\phi}_{t-1}
=
\frac{\pi}{2}
-
\frac{\pi c_y}{H}.
\label{eq:putrack_latitude}
\end{equation}

Since the spherical metric satisfies
$ds^2=\cos^2\phi\,d\lambda^2+d\phi^2$,
horizontal geometric distortion
increases toward the poles.

To maintain latitude-adaptive
search coverage, LAPS introduces
a horizontal expansion factor:

\begin{equation}
\alpha_t =
\operatorname{clip}
\left(
\frac{1}{\max(\cos\widehat{\phi}_{t-1},\epsilon)},
1,\alpha_{\max}
\right),
\label{eq:gasa_alpha}
\end{equation}

where $\epsilon=0.20$ ensures numerical
stability near the poles, and
$\alpha_{\max}=1.25$ limits the maximum
horizontal search expansion. The expansion factor is capped at
$\alpha_{\max}=1.25$ to avoid excessive
search enlargement at high latitudes. This choice is supported by Table~\ref{tab:gasa_expansion}.

For normalized search-grid coordinates
$(p,q)\in[0,1]^2$, the adapted
ERP sampling coordinates are defined as

\begin{equation}
\begin{aligned}
u_t(p)
&=c_x+\alpha_t L_t(p-\tfrac{1}{2}),\\
v_t(q)
&=c_y+L_t(q-\tfrac{1}{2}).
\end{aligned}
\label{eq:gasa_sampling}
\end{equation}

LAPS therefore expands horizontal
search coverage according to target
latitude while preserving the original
vertical search extent and fixed
network input resolution.

\begin{table*}[t!]
\centering
\setlength{\belowcaptionskip}{0pt}
\caption{
\textbf{Zero-shot tracking on PanoUAV-Track.}
Models are evaluated using their originally released pretrained
checkpoints without fine-tuning on PanoUAV-Track.
$\mathrm{AUC}_{\mathrm{C}}$ and $\mathrm{AUC}_{\mathrm{S}}$
measure full-trajectory and seam-frame robustness, respectively.
FPS reports the mean $\pm$ standard deviation over three
repetitions on a single NVIDIA H100.
Bold and underlined mean values indicate the best
and second-best results, respectively.
Red values denote the improvements over
the second-best methods.
}
\label{tab:zeroshot_tracking}
\vspace{-3mm}
\scriptsize
\setlength{\tabcolsep}{5pt}
\renewcommand{\arraystretch}{1.1}
\setlength{\arrayrulewidth}{0.4pt}
\resizebox{\linewidth}{!}{%
\begin{tabular}{l|l|l|cc|lll|ll|l}
\hline
\rowcolor{gray!12}
\multicolumn{5}{c|}{Evaluation setup} &
\multicolumn{3}{c|}{All targets (\%)} &
\multicolumn{2}{c|}{Seam robustness (\%)} & \\

\rowcolor{gray!12}
Method & Publication & Inference Checkpoint &
Visual & Language &
AUC $\uparrow$ &
$\mathrm{P}_{\mathrm{Norm}}$ $\uparrow$ &
$\mathrm{P}_{\mathrm{Ang}}$ $\uparrow$ &
$\mathrm{AUC}_{\mathrm{C}}$ $\uparrow$ &
$\mathrm{AUC}_{\mathrm{S}}$ $\uparrow$ &
FPS \\
\hline

\rowcolor{gray!8}
\multicolumn{11}{l}{\textbf{Foundation Models}} \\

SAM 2.1~\citep{ravi2025sam2}
& ICLR'25
& \texttt{sam2.1\_hiera\_large.pt}
& \cmark & \xmark
& 76.81
& \underline{83.80}
& \underline{87.00}
& 62.97
& 57.70
& $7.92 \pm 0.11$ \\

SAM2Long~\citep{ding2025sam2long}
& ICCV'25
& \texttt{sam2.1\_hiera\_large.pt}
& \cmark & \xmark
& 76.77
& 83.68
& 86.91
& 62.40
& 57.83
& $5.39 \pm 0.01$ \\

SAM 3.1~\citep{carion2026sam3}
& ICLR'26
& \texttt{sam3.1\_multiplex.pt}
& \cmark & \xmark
& 73.51
& 81.53
& 85.10
& 65.70
& 55.52
& $9.79 \pm 0.08$ \\

\hline
\rowcolor{gray!8}
\multicolumn{11}{l}{\textbf{Visual-only Trackers}} \\

OSTrack~\citep{ye2022joint}
& ECCV'22
& \texttt{OSTrack\_ep0300.pth.tar}
& \cmark & \xmark
& 75.00
& 79.80
& 83.33
& 63.81
& 57.92
& $143.74 \pm 1.71$ \\

ORTrack~\citep{wu2025ort}
& CVPR'25
& \texttt{ORTrack\_ep0300.pth.tar}
& \cmark & \xmark
& 72.74
& 77.57
& 81.69
& 62.98
& 57.19
& $\underline{214.85} \pm 12.21$ \\

SGLATrack~\citep{xue2025sglatrack}
& CVPR'25
& \texttt{sglatrack\_ep0297.pth.tar}
& \cmark & \xmark
& 73.33
& 78.05
& 82.30
& 63.11
& 57.21
& $\mathbf{234.45} \pm 6.15$ \\

ODTrack~\citep{zheng2024odtrack}
& AAAI'24
& \texttt{ODTrack\_ep0300.pth.tar}
& \cmark & \xmark
& 76.55
& 81.64
& 85.91
& 63.56
& 58.03
& $63.23 \pm 0.93$ \\

MCITrack~\citep{kang2025exploring}
& AAAI'25
& \texttt{MCITRACK\_ep0300.pth.tar}
& \cmark & \xmark
& \underline{77.63}
& 81.23
& 84.60
& \underline{66.37}
& \underline{59.76}
& $44.38 \pm 0.47$ \\

MambaLCT~\citep{li2025mambalct}
& AAAI'25
& \texttt{MambaLCT\_ep0280.pth.tar}
& \cmark & \xmark
& 75.25
& 81.06
& 85.68
& 63.37
& 57.23
& $73.76 \pm 2.25$ \\

LETrack~\citep{letrack}
& CVPR'26
& \texttt{LETrack\_ep0298.pth.tar}
& \cmark & \xmark
& 74.31
& 79.01
& 83.25
& 63.35
& 57.65
& $122.66 \pm 2.86$ \\

\hline
\rowcolor{gray!8}
\multicolumn{11}{l}{\textbf{Language-only Trackers}} \\

JointNLT~\citep{zhou2023jointnlt}
& CVPR'23
& \texttt{JointNLT\_ep0300.pth.tar}
& \xmark & \cmark
& 16.06
& 16.05
& 14.37
& 32.67
& 29.76
& $46.00 \pm 2.31$ \\

MVLM~\citep{park2026mvlm}
& CVPR'26
& \texttt{MVLM\_TF.pth.tar}
& \xmark & \cmark
& 9.42
& 9.26
& 10.06
& 9.98
& 2.99
& $83.98 \pm 2.07$ \\

\hline
\rowcolor{gray!8}
\multicolumn{11}{l}{\textbf{Vision--Language Trackers}} \\

DUTrack~\citep{li2025dutrack}
& CVPR'25
& \texttt{DUTrack\_ep0045.pth.tar}
& \cmark & \cmark
& 75.75
& 81.07
& 85.99
& 62.68
& 56.72
& $12.46 \pm 0.26$ \\

UVLTrack~\citep{ma2024uvltrack}
& AAAI'24
& \texttt{UVLTrack\_ep0300.pth.tar}
& \cmark & \cmark
& 74.80
& 79.23
& 82.99
& 64.26
& 57.91
& $65.60 \pm 0.60$ \\

UETrack~\citep{kang2026uetrack}
& CVPR'26
& \texttt{uetrack\_base.tar}
& \cmark & \cmark
& 62.75
& 67.03
& 71.44
& 61.23
& 54.66
& $154.82 \pm 9.51$ \\

\hline

\ourscell{\textcolor{darkred}{\textbf{PUTrack}}}
& \ourscell{Ours}
& \ourscell{\texttt{OSTrack\_ep0300.pth.tar}}
& \ourscell{\cmark}
& \ourscell{\xmark}
& \ourscell{\textbf{84.75}\textcolor{darkred}{+7.12}}
& \ourscell{\textbf{89.26}\textcolor{darkred}{+5.46}}
& \ourscell{\textbf{95.79}\textcolor{darkred}{+8.79}}
& \ourscell{\textbf{86.52}\textcolor{darkred}{+20.15}}
& \ourscell{\textbf{85.92}\textcolor{darkred}{+26.16}}
& \ourscell{$112.73 \pm 4.62$} \\
\hline
\end{tabular}%
}
\vspace{-3mm}
\end{table*}

\begin{table*}[t!]
\centering
\setlength{\belowcaptionskip}{0pt}

\caption{
\textbf{In-domain tracking on PanoUAV-Track.}
Trackers are trained on the PanoUAV-Track training
split and evaluated on its test split.
Pretrained encoders and auxiliary teacher models are
retained where applicable, as detailed in the appendix.
}
\label{tab:indomain_tracking}
\vspace{-2.5mm}
\scriptsize
\setlength{\tabcolsep}{5pt}
\renewcommand{\arraystretch}{1.1}
\setlength{\arrayrulewidth}{0.4pt}

\resizebox{\linewidth}{!}{%
\begin{tabular}{l|l|l|cc|lll|ll|l}
\hline
\rowcolor{gray!12}
\multicolumn{5}{c|}{Evaluation setup}
& \multicolumn{3}{c|}{All targets (\%)}
& \multicolumn{2}{c|}{Seam robustness (\%)}
& \\

\rowcolor{gray!12}
Method & Publication & Backbone
& Visual & Language
& AUC $\uparrow$
& $\mathrm{P}_{\mathrm{Norm}}$ $\uparrow$
& $\mathrm{P}_{\mathrm{Ang}}$ $\uparrow$
& $\mathrm{AUC}_{\mathrm{C}}$ $\uparrow$
& $\mathrm{AUC}_{\mathrm{S}}$ $\uparrow$
& FPS \\
\hline

\rowcolor{gray!8}
\multicolumn{11}{l}{\textbf{Visual-only Trackers}} \\

OSTrack~\citep{ye2022joint}
& ECCV'22 & ViT-B/16
& \cmark & \xmark
& 77.15 & 81.54 & 84.99
& 62.28 & 56.34
& $143.74 \pm 1.71$ \\

ORTrack~\citep{wu2025ort}
& CVPR'25 & DeiT-T/16
& \cmark & \xmark
& 74.06 & 78.96 & 82.81
& 59.69 & 54.04
& $\underline{214.85} \pm 12.21$ \\

SGLATrack~\citep{xue2025sglatrack}
& CVPR'25 & DeiT-T/16-Distilled
& \cmark & \xmark
& 75.49 & 80.24 & 83.75
& 60.48 & 54.73
& $\mathbf{234.45} \pm 6.15$ \\

\hline
\rowcolor{gray!8}
\multicolumn{11}{l}{\textbf{Language-only Trackers}} \\

JointNLT~\citep{zhou2023jointnlt}
& CVPR'23 & Swin-B/4
& \xmark & \cmark
& 51.85 & 55.07 & 56.11
& 48.35 & 41.36
& $46.00 \pm 2.31$ \\

MVLM~\citep{park2026mvlm}
& CVPR'26 & Fast-iTPN-B/16
& \xmark & \cmark
& 8.14 & 7.30 & 7.50
& 11.67 & 5.31
& $83.98 \pm 2.07$ \\

\hline
\rowcolor{gray!8}
\multicolumn{11}{l}{\textbf{Vision--Language Trackers}} \\

DUTrack~\citep{li2025dutrack}
& CVPR'25 & Fast-iTPN-B/16
& \cmark & \cmark
& 76.61 & 80.85 & 85.16
& 62.98 & 56.94
& $12.46 \pm 0.26$ \\

UVLTrack~\citep{ma2024uvltrack}
& AAAI'24 & ViT-B/16
& \cmark & \cmark
& \underline{77.49}
& \underline{81.73}
& \underline{85.18}
& \underline{63.37}
& \underline{57.31}
& $65.60 \pm 0.60$ \\

UETrack~\citep{kang2026uetrack}
& CVPR'26 & Fast-iTPN-T/16-6L
& \cmark & \cmark
& 76.23 & 81.42 & 84.73
& 61.67 & 55.98
& $154.82 \pm 9.51$ \\

\hline
\ourscell{\textcolor{darkred}{\textbf{PUTrack}}}
& \ourscell{Ours}
& \ourscell{ViT-B/16}
& \ourscell{\cmark}
& \ourscell{\xmark}
& \ourscell{\textbf{86.57}\textcolor{darkred}{+9.08}}
& \ourscell{\textbf{91.14}\textcolor{darkred}{+9.41}}
& \ourscell{\textbf{97.70}\textcolor{darkred}{+12.52}}
& \ourscell{\textbf{85.64}\textcolor{darkred}{+22.27}}
& \ourscell{\textbf{84.48}\textcolor{darkred}{+27.17}}
& \ourscell{$112.73 \pm 4.62$} \\

\hline
\end{tabular}%
}
\vspace{-3mm}
\end{table*}


\subsection{Seam-Aware Search Cropping}
\label{sec:sasc}
\indent

As shown in \cref{fig:gasa_method}(b),
the left and right boundaries of ERP
correspond to adjacent spherical directions.
Conventional planar cropping ignores
this periodic structure, potentially
fragmenting targets crossing the ERP seam.

SASC addresses this discontinuity
through horizontal periodic sampling.
Using the adapted coordinates from
Eq.~\eqref{eq:gasa_sampling},
the search image is constructed as

\begin{equation}
S_t(p,q)=
\mathcal{B}_{\mathrm{per}}
\left(
I_t,\,
u_t(p)\bmod W,\,
v_t(q)
\right),
\label{eq:gasa_crop}
\end{equation}

where $\mathcal{B}_{\mathrm{per}}$
denotes bilinear interpolation with
horizontal periodic boundary handling.

When the sampling region extends beyond
either horizontal boundary, pixels are
sampled from the opposite side of the
ERP image.
This reconstructs a spatially continuous
search observation for seam-crossing targets.


\subsection{Geometry-Consistent Coordinate Recovery}
\label{sec:recovery}
\indent

Since PUTrack constructs search regions
using latitude-adaptive sampling,
network predictions must be mapped
back into the original ERP coordinates
to preserve geometric consistency.

Let
$\widehat{B}_t^{s}
=(\widehat{p}_t,\widehat{q}_t,
\widehat{w}_t^{s},\widehat{h}_t^{s})$
denote the predicted bounding box
in normalized search coordinates,
where $(\widehat{p}_t,\widehat{q}_t)$
represents its top-left corner.

The corresponding ERP bounding box
$\widehat{B}_t
=(\widehat{x}_t,\widehat{y}_t,
\widehat{w}_t,\widehat{h}_t)$
is recovered through

\begin{equation}
\begin{aligned}
\widehat{x}_t
&=u_t(\widehat{p}_t)\bmod W,\\
\widehat{y}_t
&=v_t(\widehat{q}_t),\\
\widehat{w}_t
&=\alpha_t L_t\widehat{w}_t^{s},\\
\widehat{h}_t
&=L_t\widehat{h}_t^{s}.
\end{aligned}
\label{eq:gasa_recovery}
\end{equation}

The horizontal position is normalized
periodically, while the predicted box
width is preserved without truncation.
Therefore, seam-crossing targets retain
a geometrically consistent bounding-box
representation across ERP boundaries.

The recovered prediction $\widehat{B}_t$
is subsequently used to construct
the search region at frame $t+1$.
Through the coordinated use of LAPS,
SASC, and geometry-consistent recovery,
PUTrack enables continuous full-spherical
target tracking without ground-truth
updates or additional network
forward passes.
\section{Experiments}
\label{sec:experiments}

\subsection{Experimental Setup}
\indent

\textbf{Data Split.}
We partition PanoUAV-Track into training, validation,
and test sets following a scene-disjoint
protocol.
The resulting splits contain 1,160/166
/569 sequences, 1,799/277/796 annotated targets, and
327,091/47,100/133,770 frames for training, validation, and
testing, respectively.
Each subset includes daytime, nighttime, and simulated
sequences to ensure coverage of diverse observation
conditions. 

\textbf{Implementation Details.}
All experiments are conducted on a server equipped
with eight NVIDIA H100 GPUs.
For fair inference-speed comparisons, we benchmark
each tracker separately on a single NVIDIA H100 GPU
under a unified evaluation protocol.
Each measurement is repeated three times, and
we report the mean FPS and standard deviation.
FPS measurements include tracking inference and
preprocessing but exclude image decoding and
first-frame initialization.

\textbf{Evaluation Metrics.}
We evaluate visual-prompt and joint visual--language
tracking using Success AUC, Normalized Precision
($P_{\mathrm{Norm}}$), and Angular Precision
($P_{\mathrm{Ang}}$).
All tracking metrics follow the evaluation protocol
described in Sec.~\ref{sec:evaluation}, with further
details provided in the appendix.
  \begin{table}[t!]
  \centering
  \scriptsize
  \setlength{\tabcolsep}{2pt}
  \renewcommand{\arraystretch}{1.08}
  \setlength{\arrayrulewidth}{0.4pt}

  \caption{
  \textbf{Zero-Shot Ablation Analysis of GASA.}
  (a) Component ablation.
  (b) Horizontal search expansion.
  (c) Cropping strategies and complete framework adapters.
  }
  \label{tab:gasa_analysis}

  \vspace{-3mm}


  \begin{subtable}{\linewidth}
  \centering
  \scriptsize
  \caption{Component ablation.}
  \label{tab:gasa_components}
  \vspace{-0.5mm}

  \begin{tabularx}{\linewidth}{
  @{}>{\raggedright\arraybackslash}X|cc|cccc@{}
  }
  \hline
  \rowcolor{gray!12}
  Method &
  LAPS &
  SASC &
  AUC $\uparrow$ &
  AUCC $\uparrow$ &
  AUCS $\uparrow$ &
  FPS \\
  \hline

  OSTrack
  & $\times$ & $\times$
  & 75.00 & 63.81 & 57.92 & 143.74 \\

  + LAPS
  & $\checkmark$ & $\times$
  & 75.63 & 64.46 & 58.20 & 121.70 \\

  + SASC
  & $\times$ & $\checkmark$
  & 84.06 & 86.35 & 85.58 & 120.91 \\

  \ourscell{\textbf{+ Full GASA}}
  & \ourscell{$\checkmark$}
  & \ourscell{$\checkmark$}
  & \ourscell{\textbf{84.75}}
  & \ourscell{\textbf{86.52}}
  & \ourscell{\textbf{85.92}}
  & \ourscell{112.73} \\

  \hline
  \end{tabularx}
  \end{subtable}

  \vspace{0mm}


  \begin{subtable}{\linewidth}
  \centering
  \scriptsize
  \caption{Horizontal search expansion analysis.}
  \label{tab:gasa_expansion}
  \vspace{-0.5mm}

  \begin{tabularx}{\linewidth}{
  @{}>{\raggedright\arraybackslash}X|l|cccc@{}
  }
  \hline
  \rowcolor{gray!12}
  Strategy &
  Parameter &
  Overall $\uparrow$ &
  Equator $\uparrow$ &
  Middle $\uparrow$ &
  Polar $\uparrow$ \\
  \hline

  Fixed & $\alpha=1.00$
  & 84.06 & 83.45 & 85.75 & 79.37 \\

  Fixed & $\alpha=1.10$
  & 84.38 & 83.14 & 86.02 & 80.28 \\

  Fixed & $\alpha=1.20$
  & 84.57 & 82.97 & \textbf{86.31} & 81.22 \\

  Fixed & $\alpha=1.25$
  & 84.41 & 82.77 & 86.10 & 80.65 \\

  Fixed & $\alpha=1.30$
  & 84.38 & 82.89 & 86.13 & 81.12 \\

  Fixed & $\alpha=1.40$
  & 83.73 & 79.72 & 85.46 & \textbf{81.54} \\

  Fixed & $\alpha=1.50$
  & 82.93 & 77.95 & 84.95 & 81.38 \\

  \hline

  \ourscell{Lat.-adaptive}
  & \ourscell{$\alpha_{\max}=1.25$}
  & \ourscell{\textbf{84.75}}
  & \ourscell{\textbf{83.82}}
  & \ourscell{86.20}
  & \ourscell{80.77} \\

  Lat.-adaptive & $\alpha_{\max}=1.50$
  & 84.41 & 82.63 & 85.63 & 81.47 \\

  Lat.-adaptive & $\alpha_{\max}=1.75$
  & 84.13 & 82.44 & 85.34 & 80.64 \\

  Lat.-adaptive & $\alpha_{\max}=2.00$
  & 83.43 & 82.15 & 84.64 & 79.52 \\

  \hline
  \end{tabularx}
  \end{subtable}

  \vspace{0mm}


  \begin{subtable}{\linewidth}
  \centering
  \scriptsize
  \caption{Cropping strategies and framework adapters.}
  \label{tab:seam_crop}
  \vspace{-0.5mm}

  \setlength{\tabcolsep}{2pt}

  \resizebox{\linewidth}{!}{%
  \begin{tabular}{@{}c|l|c|cc|ccc|c@{}}
  \hline

  \rowcolor{gray!12}
  ID &
  Strategy / Adapter &
  Fwd. &
  AUC $\uparrow$ &
  AUCC $\uparrow$ &
  \multicolumn{3}{c|}{\cellcolor{gray!12}AUCS $\uparrow$} &
  FPS \\

  \rowcolor{gray!12}
  & & & & &
  All & Real & Sim. & \\

  \hline

  S0 & Constant Padding
  & 1 & 75.63 & 64.46 & 58.20 & 40.04 & 64.45 & 121.70 \\

  S1 & Border Replication
  & 1 & 75.57 & 64.05 & 57.06 & 38.18 & 63.57 & 107.93 \\

  S2 & Reflection Padding
  & 1 & 74.31 & 61.18 & 53.23 & 37.26 & 58.74 & 105.66 \\

  S3 & Periodic Nearest
  & 1 & 83.41 & 85.55 & 85.35 & 74.97 & 88.93 & 113.84 \\

  \ourscell{\textbf{S4}}
  & \ourscell{\textbf{Periodic Bilinear}}
  & \ourscell{1}
  & \ourscell{\textbf{84.75}}
  & \ourscell{\textbf{86.52}}
  & \ourscell{\textbf{85.92}}
  & \ourscell{\textbf{75.28}}
  & \ourscell{\textbf{89.58}}
  & \ourscell{112.73} \\

  S5 & BFoV/eBFoV
  & 1 & 65.57 & 75.22 & 73.41 & 39.91 & 84.96 & 83.22 \\

  S6 & Dual-view Search
  & 1--2 & 84.55 & 86.44 & 85.42 & 74.48 & 89.19 & 103.09 \\

  \hline

  F0 & 360VOT adapter
  & 1 & 77.38 & 75.10 & 74.58 & 68.62 & 76.63 & 82.21 \\

  F1 & 360VOTS adapter
  & 1 & 66.85 & 71.08 & 68.05 & 48.64 & 74.74 & 80.06 \\

  \hline
  \end{tabular}%
  }

  \end{subtable}

  \vspace{0.5mm}


\parbox{\linewidth}{
\scriptsize
\textit{Notes.}
AUCS is the overall seam accuracy; Real AUCS and Sim. AUCS
are computed separately for real and simulated targets.
S0--S6 share the same LAPS configuration; F0--F use independent
spherical search policies.
LAPS uses $\epsilon=0.20$ and
$\alpha_{\max}=1.25$ in (a) and (c).
SASC denotes periodic bilinear search sampling.
All variants in (b) enable SASC.
}

\vspace{-4.3mm}

\end{table}

\subsection{Benchmarking Existing Trackers}
\indent

We evaluate representative foundation models,
visual-only, language-only, and vision--language
trackers on PanoUAV-Track under zero-shot and
in-domain settings, as reported in
Tables~\ref{tab:zeroshot_tracking}
and~\ref{tab:indomain_tracking}.

\textbf{Zero-shot Evaluation.}
Table~\ref{tab:zeroshot_tracking} reports results
using officially released checkpoints without
fine-tuning on PanoUAV-Track.
Among visual-only trackers, MCITrack achieves
the highest AUC (77.63\%) and seam-frame AUC
(59.76\%).
SAM 2.1 reaches 76.81\% AUC, while DUTrack
achieves 75.75\%, showing competitive performance
from foundation and vision--language models.
In contrast, language-only trackers perform
poorly, with JointNLT and MVLM obtaining
16.06\% and 9.42\% AUC, respectively.
Most visual and vision--language trackers
show lower accuracy near ERP
boundaries, indicating limited robustness
to seam crossings.

\textbf{In-domain Evaluation.}
As shown in Table~\ref{tab:indomain_tracking},
training on PanoUAV-Track generally improves
tracking accuracy.
OSTrack increases from 75.00\% to 77.15\% AUC,
while ORTrack and SGLATrack gain 1.32 and
2.16 points, respectively.
UVLTrack achieves the highest baseline AUC
of 77.49\%, and UETrack improves from
62.75\% to 76.23\%.
Language-only trackers show mixed results:
JointNLT reaches 51.85\%, whereas MVLM
remains below 10\%.
Despite these improvements, seam-crossing
targets remain challenging.
OSTrack achieves 77.15\% overall AUC but
only 56.34\% seam-frame AUC, with similar
degradation observed for UVLTrack.
This suggests that in-domain training alone
is insufficient to handle ERP boundary
crossings, motivating the geometry-aware
design of PUTrack.

\subsection{Performance Comparison of PUTrack}
\label{sec:gasa_effectiveness}
\indent

\textbf{Zero-shot Evaluation.}
As shown in Table~\ref{tab:zeroshot_tracking},
PUTrack achieves the highest AUC of 84.75\%
among the methods evaluated in the main comparison,
outperforming MCITrack by 7.12 percentage points.
For seam-crossing targets, PUTrack achieves
86.52\% $\mathrm{AUC}_{\mathrm{C}}$ and
85.92\% $\mathrm{AUC}_{\mathrm{S}}$,
exceeding the best baselines by 20.15 and
26.16 points, respectively.
Compared with OSTrack using the same pretrained
checkpoint, PUTrack improves AUC by 9.75 points
while maintaining 112.73 FPS.

\textbf{In-domain Evaluation.}
Table~\ref{tab:indomain_tracking} shows that
PUTrack maintains its advantage after training
on PanoUAV-Track.
It achieves 86.57\% AUC, surpassing the
second-best UVLTrack by 9.08 points.
The improvements are more pronounced for
seam-crossing targets, with
$\mathrm{AUC}_{\mathrm{C}}$ and
$\mathrm{AUC}_{\mathrm{S}}$ exceeding
UVLTrack by 22.27 and 27.17 points.
Compared with the in-domain OSTrack baseline,
PUTrack improves AUC by 9.42 points
and operates at 112.73 FPS.
These results demonstrate the effectiveness
of PUTrack, particularly near ERP boundaries.

\subsection{Further Analysis of GASA}
\indent

We show the contributions of LAPS and SASC,
search expansion, and seam-aware cropping strategies
in Table~\ref{tab:gasa_analysis}.

\textbf{Component Ablation.}
Table~\ref{tab:gasa_components} shows that LAPS
improves baseline AUC from 75.00\% to 75.63\%,
while SASC provides a larger gain of 9.06 percentage points.
Combining both achieves the best AUC of 84.75\%,
exceeding the baseline by 9.75 points.
Full GASA also improves $\mathrm{AUC}_{\mathrm{C}}$
and $\mathrm{AUC}_{\mathrm{S}}$ from 63.81\% and
57.92\% to 86.52\% and 85.92\%, respectively,
highlighting the contribution of seam-aware sampling
to full-spherical tracking.

\providecolor{lapsgreen}{RGB}{0,85,55}

\begin{table}[t]
\centering

\caption{
\textbf{Zero-shot perspective-to-spherical generalization on PanoUAV-Track.}
$\Delta_{\mathrm{geo}} =
\mathrm{AUC}_{\mathrm{ERP}} -
\mathrm{AUC}_{\mathrm{PER}}$.
Green values indicate performance degradation
from perspective to spherical.
}
\label{tab:rq12_generalization}

\vspace{-2mm}

\scriptsize
\setlength{\tabcolsep}{2pt}
\renewcommand{\arraystretch}{1.05}
\setlength{\arrayrulewidth}{0.4pt}

\resizebox{\columnwidth}{!}{%
\begin{tabular}{l|ccc|ccc|ccc}

\noalign{\global\setlength{\arrayrulewidth}{0.6pt}}
\hline
\noalign{\global\setlength{\arrayrulewidth}{0.4pt}}

\rowcolor{gray!10}
& \multicolumn{3}{c|}{Day}
& \multicolumn{3}{c|}{Night}
& \multicolumn{3}{c}{Sim.} \\

\rowcolor{gray!10}
Method
& $\mathrm{AUC}_{\mathrm{PER}}$
& $\mathrm{AUC}_{\mathrm{ERP}}$
& $\Delta_{\mathrm{geo}}$
& $\mathrm{AUC}_{\mathrm{PER}}$
& $\mathrm{AUC}_{\mathrm{ERP}}$
& $\Delta_{\mathrm{geo}}$
& $\mathrm{AUC}_{\mathrm{PER}}$
& $\mathrm{AUC}_{\mathrm{ERP}}$
& $\Delta_{\mathrm{geo}}$ \\

\hline

OSTrack~\citep{ye2022joint}
& 85.73 & 78.13 & \textcolor{lapsgreen}{-7.60}
& 85.09 & 72.62 & \textcolor{lapsgreen}{-12.47}
& 92.10 & 71.75 & \textcolor{lapsgreen}{-20.35} \\

ORTrack~\citep{wu2025ort}
& 83.97 & 76.66 & \textcolor{lapsgreen}{-7.31}
& 82.22 & 66.38 & \textcolor{lapsgreen}{-15.84}
& 90.67 & 70.65 & \textcolor{lapsgreen}{-20.02} \\

ODTrack~\citep{zheng2024odtrack}
& 84.62 & 78.25 & \textcolor{lapsgreen}{-6.37}
& 85.96 & 81.48 & \textcolor{lapsgreen}{-4.48}
& 91.68 & 71.14 & \textcolor{lapsgreen}{-20.54} \\

MCITrack~\citep{kang2025exploring}
& 86.03 & 79.76 & \textcolor{lapsgreen}{-6.27}
& 86.37 & 78.13 & \textcolor{lapsgreen}{-8.24}
& 93.58 & 74.18 & \textcolor{lapsgreen}{-19.40} \\

UVLTrack~\citep{ma2024uvltrack}
& 84.57 & 77.63 & \textcolor{lapsgreen}{-6.94}
& 83.02 & 72.97 & \textcolor{lapsgreen}{-10.05}
& 92.11 & 71.67 & \textcolor{lapsgreen}{-20.44} \\

DUTrack~\citep{li2025dutrack}
& 83.42 & 77.52 & \textcolor{lapsgreen}{-5.90}
& 84.94 & 81.09 & \textcolor{lapsgreen}{-3.85}
& 89.61 & 70.00 & \textcolor{lapsgreen}{-19.61} \\

\ourscell{\textbf{PUTrack (Ours)}}
& \ourscell{\textbf{85.60}} & \ourscell{\textbf{84.75}} & \ourscell{\textcolor{lapsgreen}{\textbf{-0.85}}}
& \ourscell{\textbf{85.30}} & \ourscell{\textbf{77.02}} & \ourscell{\textcolor{lapsgreen}{\textbf{-8.28}}}
& \ourscell{\textbf{92.01}} & \ourscell{\textbf{89.26}} & \ourscell{\textcolor{lapsgreen}{\textbf{-2.75}}} \\

\noalign{\global\setlength{\arrayrulewidth}{0.6pt}}
\hline
\noalign{\global\setlength{\arrayrulewidth}{0.4pt}}

\end{tabular}%
}

\vspace{-1mm}
\end{table}

\begin{table}[t]
\centering
\definecolor{darkred}{RGB}{160,30,30}
\definecolor{darkgreen}{RGB}{0,100,50}

\caption{
\textbf{Zero-shot geometry-aware analysis on PanoUAV-Track.}
All scores are AUC (\%) without fine-tuning.
Equator, Middle, and Polar correspond to absolute latitudes
$<30^\circ$, $[30^\circ,60^\circ)$, and $\geq60^\circ$.
Seam denotes $\pm15$-frame windows around GT-center jumps
$|c_{x,t}-c_{x,t-1}|>W/2$ and differs from
the metric $\mathrm{AUC}_{\mathrm{S}}$.
}

\label{tab:rq3_geometry}
\vspace{-2mm}
\scriptsize
\setlength{\tabcolsep}{3pt}
\renewcommand{\arraystretch}{1.05}
\setlength{\arrayrulewidth}{0.4pt}
\resizebox{\columnwidth}{!}{%
\begin{tabular}{l|c|ccccc|ccc}
\noalign{\global\setlength{\arrayrulewidth}{0.6pt}}
\hline
\noalign{\global\setlength{\arrayrulewidth}{0.4pt}}
\rowcolor{gray!10}
&  & \multicolumn{5}{c|}{Latitude Distortion} & \multicolumn{3}{c}{Seam Discontinuity} \\
\rowcolor{gray!10}
Method & Overall  & Equator & Middle & Polar & \shortstack{M$-$E} & \shortstack{P$-$M} & Non-Seam & Seam & Drop \\
\hline
\rowcolor{gray!8}
\multicolumn{10}{l}{\textbf{Foundation Models}} \\

SAM 2.1~\citep{ravi2025sam2}
& 76.81
& 72.52 & 79.96 & 79.87 & \textcolor{darkred}{+7.44} & \textcolor{darkgreen}{-0.09}
& 77.91 & 38.21 & \textcolor{darkgreen}{-39.70} \\

SAM2Long~\citep{ding2025sam2long}
& 76.77
& 70.46 & 79.41 & 80.41 & \textcolor{darkred}{+8.95} & \textcolor{darkred}{+1.00}
& 76.76 & 38.74 & \textcolor{darkgreen}{-38.02} \\

SAM 3.1~\citep{carion2026sam3}
& 73.51
& 71.86 & 76.91 & 79.79 & \textcolor{darkred}{+5.05} & \textcolor{darkred}{+2.88}
& 75.98 & 44.22 & \textcolor{darkgreen}{-31.76} \\

\hline

\rowcolor{gray!8}
\multicolumn{10}{l}{\textbf{Visual-only Trackers}} \\

OSTrack~\citep{ye2022joint}
& 75.00
& 72.04 & 78.61 & 73.28 & \textcolor{darkred}{+6.57} & \textcolor{darkgreen}{-5.33}
& 76.47 & 34.68 & \textcolor{darkgreen}{-41.79} \\

ORTrack~\citep{wu2025ort}
& 72.74
& 67.41 & 76.44 & 66.96 & \textcolor{darkred}{+9.03} & \textcolor{darkgreen}{-9.48}
& 72.66 & 34.63 & \textcolor{darkgreen}{-38.03} \\

SGLATrack~\citep{xue2025sglatrack}
& 73.33
& 67.41 & 76.64 & 68.05 & \textcolor{darkred}{+9.23} & \textcolor{darkgreen}{-8.59}
& 72.86 & 34.89 & \textcolor{darkgreen}{-37.97} \\

ODTrack~\citep{zheng2024odtrack}
& 76.55
& 71.81 & 78.77 & 77.94 & \textcolor{darkred}{+6.96} & \textcolor{darkgreen}{-0.83}
& 76.90 & 35.82 & \textcolor{darkgreen}{-41.08} \\

MCITrack~\citep{kang2025exploring}
& 77.63
& 75.79 & 80.50 & 76.61 & \textcolor{darkred}{+4.71} & \textcolor{darkgreen}{-3.89}
& 79.31 & 37.86 & \textcolor{darkgreen}{-41.45} \\

MambaLCT~\citep{li2025mambalct}
& 75.25
& 72.67 & 78.41 & 74.94 & \textcolor{darkred}{+5.74} & \textcolor{darkgreen}{-3.47}
& 76.79 & 36.10 & \textcolor{darkgreen}{-40.69} \\

\hline

\rowcolor{gray!8}
\multicolumn{10}{l}{\textbf{Language-only Trackers}} \\

JointNLT~\citep{zhou2023jointnlt}
& 16.06
& 22.27 & 7.35 & 6.43 & \textcolor{darkgreen}{-14.92} & \textcolor{darkgreen}{-0.92}
& 13.87 & 12.62 & \textcolor{darkgreen}{-1.25} \\

MVLM~\citep{park2026mvlm}
& 9.42
& 6.85 & 11.06 & 12.82 & \textcolor{darkred}{+4.21} & \textcolor{darkred}{+1.76}
& 9.58 & 3.40 & \textcolor{darkgreen}{-6.18} \\

\hline

\rowcolor{gray!8}
\multicolumn{10}{l}{\textbf{Vision--Language Trackers}} \\

DUTrack~\citep{li2025dutrack}
& 75.75
& 71.15 & 78.12 & 77.76 & \textcolor{darkred}{+6.97} & \textcolor{darkgreen}{-0.36}
& 76.26 & 36.00 & \textcolor{darkgreen}{-40.26} \\

UVLTrack~\citep{ma2024uvltrack}
& 74.80
& 71.25 & 78.18 & 70.55 & \textcolor{darkred}{+6.93} & \textcolor{darkgreen}{-7.63}
& 75.50 & 37.99 & \textcolor{darkgreen}{-37.51} \\

UETrack~\citep{kang2026uetrack}
& 62.75
& 56.60 & 66.16 & 44.77 & \textcolor{darkred}{+9.56} & \textcolor{darkgreen}{-21.39}
& 60.62 & 31.91 & \textcolor{darkgreen}{-28.71} \\

\hline
\ourscell{\textbf{Average}}
& \ourscell{\textbf{65.46}}
& \ourscell{\textbf{62.15}}
& \ourscell{\textbf{67.61}}
& \ourscell{\textbf{63.58}}
& \ourscell{\textcolor{darkred}{\textbf{+5.46}}}
& \ourscell{\textcolor{darkgreen}{\textbf{-4.02}}}
& \ourscell{\textbf{65.82}}
& \ourscell{\textbf{32.65}}
& \ourscell{\textcolor{darkgreen}{\textbf{-33.17}}} \\
\hline
\hline
\ourscell{\textbf{PUTrack (Ours)}}
& \ourscell{\textbf{84.75}}
& \ourscell{\textbf{83.82}}
& \ourscell{\textbf{86.20}}
& \ourscell{\textbf{80.77}}
& \ourscell{\textcolor{darkred}{\textbf{+2.38}}}
& \ourscell{\textcolor{darkgreen}{\textbf{-5.43}}}
& \ourscell{\textbf{84.62}}
& \ourscell{\textbf{83.43}}
& \ourscell{\textcolor{darkgreen}{\textbf{-1.19}}} \\

\noalign{\global\setlength{\arrayrulewidth}{0.6pt}}
\hline
\noalign{\global\setlength{\arrayrulewidth}{0.4pt}}
\end{tabular}%
}
\vspace{-3mm}
\end{table}

\begin{figure*}[t!]
  \centering
  \captionsetup{skip=2pt}
  \includegraphics[width=\linewidth]{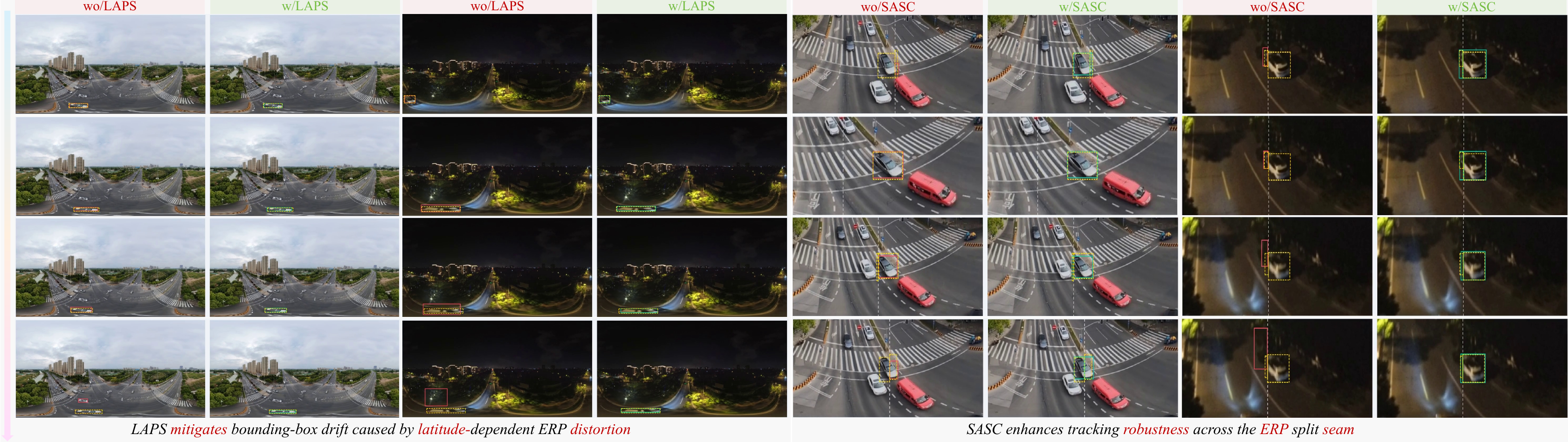}
  \caption{
  \textbf{Qualitative visualization of LAPS and SASC on PanoUAV-Track.}
  Comparisons with (w/) and without (w/o) each component
  are shown across successive frames.
  LAPS mitigates latitude-dependent drift, while SASC
  improves stability near the ERP split seam.
  }
  \label{fig:laps_sasc_visual}
  \vspace{-4mm}
\end{figure*}

\textbf{Search Expansion Analysis.}
Table~\ref{tab:gasa_expansion} compares fixed and
latitude-adaptive expansion. Among fixed strategies,
$\alpha=1.20$ achieves the highest overall AUC of 84.57\%,
while increasing $\alpha$ to 1.50 reduces it to 82.93\%.
Latitude-adaptive expansion with $\alpha_{\max}=1.25$
achieves the best overall AUC of 84.75\%.
Although $\alpha=1.40$ yields the highest polar AUC
of 81.54\%, it degrades equatorial performance,
suggesting that search context should vary with latitude.

\textbf{Seam-Aware Cropping and Framework Analysis.}
In Table~\ref{tab:seam_crop}, S0--S6 share the same LAPS
configuration, whereas F0--F1 are complete temporal adapters.
Conventional padding achieves only 74.31--75.63\% AUC
and 53.23--58.20\% $\mathrm{AUC}_{\mathrm{S}}$.
Periodic sampling improves seam handling, with S4 reaching
84.75\% AUC and 85.92\% $\mathrm{AUC}_{\mathrm{S}}$.
The F0 360VOT and F1 360VOTS adapters obtain
77.38/74.58\% and 66.85/68.05\% in
AUC/$\mathrm{AUC}_{\mathrm{S}}$, respectively.
Both remain below S4 in these metrics, extending the
comparison beyond individual cropping operators.

\textbf{Qualitative Analysis.}
\cref{fig:laps_sasc_visual} illustrates the complementary
effects of LAPS and SASC. LAPS mitigates bounding-box
drift associated with latitude-dependent ERP distortion,
improving localization under challenging spherical projections.
SASC preserves continuity across the ERP split seam,
reducing abrupt localization errors and stabilizing
tracking during seam crossings.

\textbf{Real versus Simulated Seam Robustness.}
Table~\ref{tab:seam_crop} further separates real and simulated
seam accuracy, with real data combining daytime and nighttime
sequences. Using the same full-GASA run with 84.75\% overall
AUC, S4 achieves \textbf{75.28\%}/\textbf{89.58\%}
real/simulated $\mathrm{AUC}_{\mathrm{S}}$, compared with
39.64\%/64.22\% for OSTrack.
The gains of 35.64 and 25.36 percentage points demonstrate
improvements in both domains. S4 also exceeds F0 and F1.

\subsection{Perspective-to-Spherical Analysis}
\indent

\textbf{RQ1.} \emph{How Well Do Perspective Trackers Generalize
to Full-Spherical UAV Observation?}
We evaluate generalization using matched ERP and perspective
observations, defining the gap as
$\Delta_{\mathrm{geo}}=
\mathrm{AUC}_{\mathrm{ERP}}-\mathrm{AUC}_{\mathrm{PER}}$.
Table~\ref{tab:rq12_generalization} shows that all six baselines
degrade under ERP observations. PUTrack also exhibits negative
gaps but substantially reduces daytime and simulation degradation,
indicating improved generalization to full-spherical UAV tracking.

\textbf{RQ2.} \emph{Does the Generalization Gap Persist
across Daytime, Nighttime, and Simulated Domains?}
Table~\ref{tab:rq12_generalization} reveals gaps across all
three domains. Baseline AUC degradation ranges from
5.90--7.60 points during daytime, 3.85--15.84 at night,
and 19.40--20.54 in simulation.
PUTrack reduces the corresponding gaps to 0.85, 8.28,
and 2.75 points. It also achieves the highest daytime and
simulation ERP AUC, exceeding the best baselines by 4.99
and 15.08 points, respectively. These results show
daytime and simulation improvements, while nighttime
generalization remains challenging.

\textbf{RQ3.} \emph{Is Tracking Degradation Systematically
Associated with Spherical Projection Geometry?}
Table~\ref{tab:rq3_geometry} analyzes performance across
latitude regions and ERP seam crossings.
Baseline AUC increases by 5.46 points on average from
Equator to Middle, then decreases by 4.02 points toward Polar.
PUTrack achieves 84.75 overall AUC and reduces the
Equator-to-Middle variation to 2.38 points, although its
Polar AUC remains 5.43 points below Middle.
It also reduces the seam drop from the baseline average
of 33.17 points to 1.19 points, achieving 83.43 AUC within
seam-event windows.
These findings indicate that PUTrack substantially alleviates
ERP seam discontinuity, while latitude-dependent distortion
remains challenging.
\section{Conclusion}
\label{sec:conclusion}

We introduce \textbf{PanoUAV-Track}, a large-scale
benchmark for full-spherical UAV tracking across
daytime, nighttime, and simulated scenarios.
Extensive evaluations reveal a substantial
perspective-to-spherical generalization gap,
highlighting challenges from latitude-dependent
distortion and ERP seam crossings.
To address this gap, we propose
\textbf{PanoUAV-Tracker (PUTrack)}, a geometry-aware
tracking framework incorporating the
Geometry-Aware Search Adapter (GASA).
By combining latitude-adaptive search and
seam-aware sampling, PUTrack achieves robust full-spherical
tracking without modifying the underlying network.
Experiments demonstrate consistent improvements
under both zero-shot and in-domain settings,
particularly for seam-crossing targets.

\textbf{Future Work.}
We plan to investigate ultra-long panoramic tracking
sequences, focusing on long-term tracking robustness
under full-spherical observations, including
accumulated drift and target re-acquisition.

\section*{LLM Usage Statement}

Qwen3 was used to generate initial target descriptions
for subsequent author review and correction.
ChatGPT was used to assist with language polishing
and manuscript formatting.
GPT-6 was used to assist dataset integrity
auditing.
The authors reviewed and verified all resulting annotations
and text and take full responsibility for the final
dataset and manuscript.

{
    \small
    \bibliographystyle{ieeenat_fullname}
    \bibliography{main}
}

\clearpage
\setcounter{page}{1}
\maketitlesupplementary
\appendix

\newpage
\appendix

\section{Cross-Dataset Generalization}
\label{app:cross-dataset-generalization}

To assess cross-dataset generalization, we evaluate OSTrack and PUTrack on 360Tracking, comprising 21 sequences and 9,909 annotated frames. Both methods share identical checkpoints and initialization conditions within each comparison, without dataset-specific training or hyperparameter tuning. Following our circular-IoU protocol, evaluation covers 9,725 valid visible frames, including 383 seam frames from 13 sequences. As shown in Table~\ref{tab:cross-dataset-360tracking}, PUTrack improves AUC and AUCS from 54.41\% and 32.31\% to 72.23\% and 67.18\%, respectively, using the released weights. With weights trained on our dataset, PUTrack achieves consistent improvements of 11.18 and 29.31 percentage points in AUC and AUCS, respectively. These results support the transferability of the proposed seam-aware search mechanism, particularly under seam-crossing conditions. However, the lower absolute performance of the checkpoints trained on our dataset and the limited benchmark scale warrant caution when drawing broader conclusions about cross-dataset generalization.

\section{Data Audit}

\subsection{Data Acquisition and Scene-Disjoint Split}
\label{app:data_acquisition}

The real-world videos were independently collected using
\emph{Antigravity A1} and \emph{DJI Avata 360}.
All videos were exported at 30 fps.
The original ERP frames were recorded at 8K resolution and
subsequently resized to $1920\times960$ for annotation and evaluation.
Each recording session captured a different real-world scene.
We therefore identify scenes \textbf{using their recording timestamps and
perform dataset splitting at the scene level}.
All clips and target trajectories collected at the same recording
time are assigned to the same subset, ensuring that the training,
validation, and test scenes are mutually disjoint.

\subsection{Core Data Audit and Conclusions}
\label{app:core_data_audit}

\paragraph{How we checked the dataset split.}
We read every sequence listed in the train, validation, and test files.
For each sequence, we counted its image frames and annotated targets and
checked that its image and annotation folders exist.  The current
release contains 1,895 sequences, 507,961 image frames, and 2,872 target
trajectories.

We then checked whether clips from the same recording had been placed in
different splits.  For real videos, clips with the same recording
timestamp were treated as one group.  Suffixes used for short clips and
key-frame clips were removed before this comparison.  If a full
timestamp was not available, we used the original clip name.  For
simulation data, clips from the same source episode were treated as one
group.  This produced 1,046 groups: 459 day, 298 night, and 289
simulation groups.  We found no group in more than one split.
Table~\ref{tab:core_split_inventory} gives the full counts.

The split check shows that clips from the same known recording or
simulation episode do not appear in different splits.  It also confirms
that the day, night, and simulation test sets all contain seam cases.
The domain-wise seam results are therefore calculated from actual test
samples.

\paragraph{Latitude and seam coverage.}
We divide visible post-initialization GT frames into equatorial
($|\phi|<30^\circ$), mid-latitude
($30^\circ\leq|\phi|<60^\circ$), and polar
($|\phi|\geq60^\circ$) strata.  A seam frame contains two visible GT
segments at the ERP boundary.  Table~\ref{tab:core_geometry_support}
shows that all three domains contain seam cases, although their natural
distributions differ substantially.  In particular, simulation contains
many equatorial seam frames but relatively few polar frames; the
balanced re-annotation audit must therefore be reported separately from
the natural test distribution.

\begin{table}[t]
\centering
\caption{Cross-dataset evaluation on 360Tracking (21 sequences). Each paired comparison uses identical OSTrack checkpoints without dataset-specific adaptation. All metrics are reported as percentages under our circular-box evaluation protocol.}
\label{tab:cross-dataset-360tracking}
\scriptsize
\setlength{\tabcolsep}{3pt}
\renewcommand{\arraystretch}{1.05}
\begin{tabularx}{\linewidth}{@{}lXrrr@{}}
\hline
Weights & Method & AUC$\uparrow$ & AUCC$\uparrow$ & AUCS$\uparrow$ \\
\hline
Released & OSTrack & 54.41 & 40.64 & 32.31 \\
\ourscell{Released} & \ourscell{PUTrack} & \ourscell{\textbf{72.23}} & \ourscell{\textbf{73.72}} & \ourscell{\textbf{67.18}} \\
\hline
PanoUAV-trained & OSTrack & 46.11 & 34.51 & 23.08 \\
\ourscell{PanoUAV-trained} & \ourscell{PUTrack} & \ourscell{\textbf{57.29}} & \ourscell{\textbf{54.34}} & \ourscell{\textbf{52.39}} \\
\hline
\end{tabularx}
\end{table}

\begin{table}[t]
\centering
\caption{Audited inventory of the current \texttt{final\_V2} split.
Frames are counted once per sequence, whereas targets are independent
annotated trajectories.}
\label{tab:core_split_inventory}
\small
\setlength{\tabcolsep}{4pt}
\begin{tabular}{llrrrr}
\hline
Split & Domain & Groups & Sequences & Frames & Targets \\
\hline
Train & Day   & 199 & 666 & 207,071 & 1,148 \\
      & Night & 154 & 424 &  78,827 &   581 \\
      & Sim.  &  21 &  70 &  41,193 &    70 \\
Val.  & Day   &  73 &  95 &  28,518 &   181 \\
      & Night &  34 &  61 &  10,264 &    86 \\
      & Sim.  &  10 &  10 &   8,318 &    10 \\
Test  & Day   & 187 & 190 &  62,225 &   384 \\
      & Night & 110 & 121 &  21,783 &   154 \\
      & Sim.  & 258 & 258 &  49,762 &   258 \\
\hline
Total &       & 1,046 & 1,895 & 507,961 & 2,872 \\
\hline
\end{tabular}
\end{table}

\begin{table}[t]
\centering
\caption{Valid test frames by GT-center latitude.  Parentheses give the
number of seam-split frames within each cell.  Initialization and
invisible-GT frames are excluded.}
\label{tab:core_geometry_support}
\small
\setlength{\tabcolsep}{5pt}
\begin{tabular}{lrrr}
\hline
Domain & Equatorial & Mid-latitude & Polar \\
\hline
Day   & 36,135 (541)    & 66,440 (815)  & 14,907 (479) \\
Night &  7,165 (82)     & 13,885 (216)  &  4,497 (195) \\
Sim.  & 41,458 (11,048) &  7,264 (2,601)&    583 (40)  \\
\hline
\end{tabular}
\end{table}


\section{Geometry-Aware Evaluation Details}
\label{app:geometry_eval}

We provide additional details of the geometry-aware
evaluation protocol used in PanoUAV-Track.
Our analysis considers three complementary aspects:
latitude-dependent performance variation, ERP seam
crossings, and perspective-to-spherical generalization.
The first two are evaluated by regrouping existing
tracking predictions, while the last requires
additional inference on matched perspective views.

\subsection{Latitude-Dependent Analysis}
\label{app:latitude_analysis}

We investigate how tracking accuracy varies with
the spherical latitude of the target.
Given an ERP frame of height $H$ and a ground-truth
bounding box $B_t=(x_t,y_t,w_t,h_t)$, where
$(x_t,y_t)$ denotes its top-left corner,
the target center is defined as

\begin{equation}
\begin{aligned}
c_{x,t}
&=
\left(x_t+\frac{w_t}{2}\right)\bmod W,\\
c_{y,t}
&=
y_t+\frac{h_t}{2}.
\end{aligned}
\end{equation}
The modulo operation maps the horizontal center to
the canonical ERP range $[0,W)$.

The corresponding spherical latitude is

\begin{equation}
\phi_t
=
\frac{\pi}{2}
-
\frac{\pi c_{y,t}}{H}.
\end{equation}

Target observations are divided into three regions:

\begin{equation}
\begin{aligned}
\text{Equator}:&\quad |\phi_t|<30^\circ,\\
\text{Middle}:&\quad 30^\circ\leq|\phi_t|<60^\circ,\\
\text{Polar}:&\quad |\phi_t|\geq60^\circ.
\end{aligned}
\end{equation}

We compute circular IoU for each valid target frame
and independently evaluate Success AUC by pooling
observations within each latitude region.
The resulting scores are denoted by
$AUC_E$, $AUC_M$, and $AUC_P$.

The latitude-dependent performance changes are

\begin{equation}
\begin{aligned}
\Delta_{M-E}&=AUC_M-AUC_E,\\
\Delta_{P-M}&=AUC_P-AUC_M.
\end{aligned}
\end{equation}

Negative values indicate lower performance in the
higher-latitude region.
All results are derived from the same tracking
predictions without additional inference.
This analysis characterizes latitude-dependent
performance differences without assuming that
projection distortion is their sole cause.

\subsection{ERP Seam-Crossing Analysis}
\label{app:seam_analysis}

We further evaluate tracking performance near
the horizontal ERP boundary.
A seam-crossing event is identified when the
horizontal ground-truth target center exhibits
a displacement exceeding half the image width
between two consecutive visible frames:

\begin{equation}
|c_{x,t}-c_{x,t-1}|>\frac{W}{2},
\end{equation}

where $c_{x,t}$ denotes the horizontal coordinate
of the ground-truth bounding-box center,
and $W$ denotes the ERP image width.
The center coordinates are represented within
the canonical ERP horizontal range $[0,W)$.

For each detected crossing at frame $t_c$,
we consider a temporal neighborhood

\begin{equation}
\mathcal{W}_{\mathrm{seam}}
=
[t_c-15,t_c+15].
\end{equation}

Valid observations within the union of these
windows are assigned to the Seam subset,
while the remaining observations form the
Non-Seam subset.
Overlapping windows are counted only once.

Circular-IoU-based Success AUC is independently
computed for both subsets using frame-level pooling.
The signed seam performance change is

\begin{equation}
\Delta_{\mathrm{seam}}
=
AUC_{\mathrm{seam}}
-
AUC_{\mathrm{non}}.
\end{equation}

A negative value indicates lower tracking accuracy
near seam crossings.
This analysis uses existing ERP tracking predictions
without additional network inference.


\section{Evaluation Metrics}
\label{app:evaluation_metrics}

We summarize the evaluation metrics used in
PanoUAV-Track.
Let $\mathcal{V}$ denote the set of valid evaluation
frames for a target trajectory, excluding
initialization frames and observations with
invisible ground-truth targets.
The number of valid evaluation frames is
$T=|\mathcal{V}|$.
Missing predictions on valid target frames
are treated as tracking failures.

\paragraph{Success AUC.}
Given a ground-truth bounding box $B_t$ and
a prediction $\widehat{B}_t$, let

\begin{equation}
q_t
=
\operatorname{cIoU}(B_t,\widehat{B}_t),
\end{equation}

where $\operatorname{cIoU}$ accounts for
horizontal ERP periodicity.
For missing predictions, we set $q_t=0$.

To ensure consistent treatment of zero-overlap
predictions, the success indicator is defined as

\begin{equation}
\mathcal{I}_t(\tau)
=
\begin{cases}
\mathbb{I}[q_t>0],
& \tau=0,\\
\mathbb{I}[q_t\geq\tau],
& 0<\tau\leq1.
\end{cases}
\end{equation}

The success score at threshold $\tau$ is

\begin{equation}
S(\tau)
=
\frac{1}{T}
\sum_{t\in\mathcal{V}}
\mathcal{I}_t(\tau).
\end{equation}

The success AUC is defined as

\begin{equation}
\mathrm{AUC}
=
\int_0^1 S(\tau)\,d\tau.
\end{equation}

In implementation, the success curve is
numerically integrated over uniformly spaced
IoU thresholds.
Overall AUC is computed independently for
each target trajectory and then averaged
across targets.

\paragraph{Normalized Precision.}
Let $(c_{x,t},c_{y,t})$ and
$(\widehat{c}_{x,t},\widehat{c}_{y,t})$
denote the ground-truth and predicted
bounding-box centers, respectively.

The horizontal center distance accounts for
ERP periodicity:

\begin{equation}
\begin{aligned}
d_t^x &=
\left|
(\widehat{c}_{x,t}\bmod W)
-
(c_{x,t}\bmod W)
\right|,\\
\widetilde{d}_t^x
&=\min(d_t^x,W-d_t^x).
\end{aligned}
\end{equation}

The normalized center error is

\begin{equation}
e_{\mathrm{norm}}^t
=
\sqrt{
\left(
\frac{\widetilde{d}_t^x}{w_t}
\right)^2
+
\left(
\frac{\widehat{c}_{y,t}-c_{y,t}}{h_t}
\right)^2
},
\end{equation}

where $w_t$ and $h_t$ denote the ground-truth
bounding-box width and height, respectively.
For missing predictions, the normalized center
error is assigned $+\infty$.

The normalized precision curve is defined as

\begin{equation}
P(r)
=
\frac{1}{T}
\sum_{t\in\mathcal{V}}
\mathbb{I}
\left[
e_{\mathrm{norm}}^t\leq r
\right].
\end{equation}

Normalized Precision is obtained by integrating
this curve over the threshold interval $[0,0.5]$
and normalizing by the interval length:

\begin{equation}
P_{\mathrm{Norm}}
=
\frac{1}{0.5}
\int_0^{0.5}P(r)\,dr.
\end{equation}

\paragraph{Angular Precision.}
Angular localization error is computed from
the spherical coordinates of predicted and
ground-truth target centers.

Let $(\lambda_t,\phi_t)$ and
$(\widehat{\lambda}_t,\widehat{\phi}_t)$
denote their respective spherical longitudes
and latitudes, obtained from the corresponding
ERP center coordinates.

The great-circle angular error is

\begin{equation}
\begin{aligned}
e_{\mathrm{ang}}^t
=
\arccos\Big(
\operatorname{clip}\big(
&\sin\phi_t\sin\widehat{\phi}_t\\
&+\cos\phi_t\cos\widehat{\phi}_t
\cos(\lambda_t-\widehat{\lambda}_t),
-1,1
\big)
\Big).
\end{aligned}
\end{equation}

For missing predictions, the angular error
is assigned $+\infty$.

Angular Precision is evaluated using
a fixed threshold of $3^\circ$:

\begin{equation}
P_{\mathrm{Ang}}
=
\frac{1}{T}
\sum_{t\in\mathcal{V}}
\mathbb{I}
\left[
e_{\mathrm{ang}}^t\leq3^\circ
\right].
\end{equation}

This metric measures localization accuracy
on the viewing sphere and remains continuous
across the ERP seam.

\paragraph{Seam Robustness.}
We additionally report
$\mathrm{AUC}_{\mathrm{C}}$ and
$\mathrm{AUC}_{\mathrm{S}}$ for targets
whose ground-truth boxes cross the ERP seam.

The former evaluates their complete valid
trajectories, while the latter considers
only frames in which their ground-truth
boxes intersect the seam.
Both metrics are averaged across targets.

Unlike these target-averaged metrics,
the Seam/Non-Seam analysis in
Sec.~\ref{app:seam_analysis} uses
frame-pooled observations within
temporal windows around seam crossings.

\paragraph{Metric Computation Details.}
All reported metrics follow a unified evaluation
protocol and exclude initialization frames and
invisible ground-truth observations.
Missing predictions on valid frames contribute
failures to the corresponding metric.

\noindent\textbf{Success AUC.}
We evaluate circular-IoU-based success at
21 uniformly spaced thresholds
$\tau_k=0.05k$, $k=0,\ldots,20$.

At $\tau_0=0$, success requires strictly
positive overlap, whereas all remaining
thresholds use $q_t\geq\tau_k$.

The success curve is integrated using
the trapezoidal rule:

\begin{equation}
\mathrm{AUC}
=
\sum_{k=0}^{19}
\frac{0.05}{2}
\left[
S(\tau_k)+S(\tau_{k+1})
\right].
\end{equation}

The same numerical integration protocol
is applied to $\mathrm{AUC}_{\mathrm{C}}$
and $\mathrm{AUC}_{\mathrm{S}}$.

\noindent\textbf{Normalized Precision.}
Horizontal center displacement is computed
using the shortest periodic distance
$\widetilde{d}_x^t$ defined above.

Horizontal and vertical errors are normalized
by the width and height of the ground-truth
circular bounding box, respectively.

The resulting normalized precision curve
is evaluated at thresholds
$r_k=0.01k$, $k=0,\ldots,50$.

Normalized Precision is obtained as

\begin{equation}
P_{\mathrm{Norm}}
=
\frac{1}{0.5}
\sum_{k=0}^{49}
\frac{0.01}{2}
\left[
P(r_k)+P(r_{k+1})
\right].
\end{equation}

All reported scores are multiplied by $100$
and expressed as percentages.

\subsection{Target-Centric Perspective Evaluation}
\label{app:controlled_reprojection}

We compare tracking performance on original ERP sequences and
target-centric perspective sequences generated from the same videos.
Each pair preserves the target identity and source-frame correspondence,
allowing performance to be examined under these two observation settings.

\paragraph{Perspective sequence generation and annotation.}
For each source-visible frame, the virtual camera is oriented toward
the target center while retaining temporal lag in camera orientation and residual target motion.
A rectilinear view with horizontal and vertical fields of view of
$90^\circ$ is rendered at $480\times480$ resolution by inverse ray
mapping and bilinear interpolation. ERP sampling wraps horizontally
and clamps vertical coordinates to the valid image range.
Ground-truth bounding boxes are then annotated directly in the
rendered perspective images; they are not obtained by projecting
the ERP bounding boxes. Source annotations are used only to construct
the offline perspective sequences. During tracking, each method
receives the first-frame box annotated in its respective representation
and the language or category input required by its protocol.
No subsequent ground-truth boxes or virtual-camera parameters
are supplied to the tracker.

\paragraph{Fairness of Perspective-to-Spherical Evaluation.}
Our evaluation focuses on visual target tracking rather than active viewpoint planning. Therefore, using ground-truth target locations to guide virtual camera orientation during offline perspective sequence generation is a reasonable data construction strategy. This process primarily ensures target visibility within the limited perspective field of view while retaining temporal lag in camera orientation and residual target motion.
Importantly, ground-truth information is used exclusively for offline data generation and is not provided to trackers during inference. All evaluated trackers receive identical perspective sequences under the same evaluation protocol, ensuring fairness across methods.
The perspective $AUC_{PER}$ serves as a reference for quantifying performance degradation when transferring from conventional perspective observations to full-spherical ERP observations, not as an independent benchmark for active camera tracking. Consequently, this comparison characterizes the generalization gap between the two observation settings while maintaining consistent evaluation conditions across trackers.

\paragraph{Matched evaluation.}
For each paired trajectory, evaluation uses the intersection of
source-frame indices with valid annotations in both representations,
excluding initialization and invisible-target frames.
Perspective predictions are evaluated against the newly annotated
perspective boxes using planar IoU, while ERP predictions are evaluated
using circular IoU.
\textbf{Both metrics use the same intersection-over-union definition;
circular IoU accounts for horizontal periodicity when computing
intersection and union areas. Their scores are therefore comparable
under the same overlap-based scoring protocol.}
Success AUC is computed using trapezoidal integration over 21
thresholds from 0 to 1 and averaged equally across valid targets.

\paragraph{Interpretation.}
We report the performance difference
\begin{equation}
\Delta_{\mathrm{geo}}
=
AUC_{\mathrm{ERP}}-AUC_{\mathrm{PER}},
\end{equation}
where negative values indicate higher AUC on the target-centric
perspective sequences. This comparison measures performance under
two specified observation settings. Target centering is an explicit
property of the perspective dataset construction. It changes apparent
motion, target position, field of view, and image sampling.
Moreover, IoU-based evaluation in each representation is not
generally projection-invariant. Accordingly, the difference
characterizes these paired settings and is not attributed solely
to ERP projection distortion.

\section{Search-Cropping Strategies with LAPS}
\label{app:panoramic-baselines}

This section specifies the seven search-cropping strategies
(S0--S6) compared in Table~\ref{tab:seam_crop}.

\paragraph{Shared setup.}
All variants use the same released OSTrack checkpoint,
first-frame initialization, and constant-padded ERP template
($128\times128$, template factor 2), without additional training.
The search factor is 4, and the search input is resized to
$256\times256$. Each variant applies the same LAPS rule
($\epsilon=0.20$, $\alpha_{\max}=1.25$) to its previous prediction
to determine the search center and extent. The variants run
independently, so their predicted states may diverge over time.
S0--S5 use one network forward per frame; S6 uses one or two.
For the planar crops in S0--S4 and S6, vertical overflow is
zero padded.

\paragraph{S0: Constant padding.}
The LAPS-expanded search rectangle is cropped directly from
the ERP image. Samples outside the image are filled with zeros,
and the resulting patch is resized using bilinear interpolation.
S0 therefore serves as the LAPS-enabled planar-padding control;
it does not connect content across the horizontal ERP seam.

\paragraph{S1: Border replication.}
Horizontal out-of-bounds samples are replaced by the nearest
left or right boundary pixel, followed by bilinear resizing.
This extends the boundary appearance into the missing region
instead of introducing zeros, but does not recover scene content
from the opposite ERP boundary.

\paragraph{S2: Reflection padding.}
Horizontal out-of-bounds samples are obtained by reflecting
image content at the corresponding boundary using
\texttt{BORDER\_REFLECT\_101}, which does not repeat the boundary
pixel. The patch is then resized bilinearly. Unlike periodic
sampling, reflection creates mirrored local content.

\paragraph{S3: Periodic nearest.}
When the search rectangle crosses the horizontal seam, integer
source columns are wrapped modulo the ERP width $W$ and gathered
into a contiguous patch. Otherwise, the original constant-padded
crop is used. The entire patch is resized using nearest-neighbor
interpolation on every frame. Thus, S3 changes the resizing
policy globally, not only at seam pixels.

\paragraph{S4: Periodic bilinear.}
S4 uses the same horizontal wrapping and patch assembly as S3,
but applies bilinear resizing. For rectangles that do not cross
the seam, it retains the original constant-padded bilinear crop.
This strategy joins the actual scene content on both sides of
the ERP seam within a single search input. It is the SASC
configuration used in the selected GASA variant.

\paragraph{S5: BFoV/eBFoV reprojection.}
S5 uses the official 360VOT geometry implementation
\cite{huang2023_360vot,xu2025_360vots} to construct a local view
centered on the previous prediction, with zero roll.
For an ERP image of size $W\times H$ and LAPS-expanded search
dimensions $(w_c,h_c)$, the angular extents are
\begin{equation}
\begin{aligned}
\theta_h &= \min(360^\circ,\,360^\circ w_c/W),\\
\theta_v &= \min(180^\circ,\,180^\circ h_c/H).
\end{aligned}
\end{equation}
The released automatic rule selects a tangent view when both
extents are at most $90^\circ$, and a spherical view otherwise.
The view is sampled at $256\times256$ using the library's
nearest-neighbor remapping and boundary conventions.
The predicted local box is clipped to the local image,
inverse-projected through the stored sampling map, and converted
to a circular ERP box for the next state update.

The template remains in ERP, and the search extent follows LAPS.
S5 evaluates this reprojection operator within the host tracker,
not the complete 360VOT/360VOTS temporal framework: it does not
include their separate angular search policy, loss-state holding,
progressive expansion, or full-frame recovery.

\paragraph{S6: Dual-view search.}
S6 first evaluates the constant-padded view. If the predicted
search rectangle crosses the seam, it additionally evaluates
a periodic-bilinear view, provided that this patch contains valid
image content and differs from the first patch. Both views use
the same pre-update state, template, and model weights.

Each candidate is considered supported when at least 50\% of
its predicted box is covered by valid image pixels; padding and
out-of-view area count as unsupported. If only one candidate is
supported, it is selected. If both are supported, selection uses
the peak-to-sidelobe ratio (PSR) of their Hann-windowed response
maps, excluding a $5\times5$ neighborhood around each peak.
The periodic candidate is selected only when its PSR is higher.
Ties and cases with two unsupported candidates retain the
constant-padded candidate. The tracker updates its state once
from the selected prediction, without averaging candidates or
using ground truth.

\paragraph{F0--F1: Complete framework adapters.}
F0 (360VOT) and F1 (360VOTS) use exactly the same released
OSTrack ViT-B/16 epoch-300 checkpoint as GASA, verified by
SHA-256, without fine-tuning. Both project the template and
search region to local BFoV/eBFoV views and inverse-project
predictions to update the spherical state. F0 fixes the
template/search factors to 2/4 without validation-set tuning
and accepts every finite prediction. F1 fixes
$\mathrm{SR\ Ratio}=2$ and $\mathrm{Max\ Loss}=4$, and selects
$\mathrm{SR\ Min}=5^\circ$ and confidence threshold $\tau=0.25$
from $\{5^\circ,90^\circ\}\times\{0,0.25,0.5,0.75\}$ using
target-averaged AUC on all 277 validation targets; these settings
are frozen before testing. Rejected F1 predictions count as
failures and trigger search-region holding, progressive
expansion, and ultimately full-ERP recovery. Projected inputs
may differ in appearance and target scale from OSTrack's
perspective training crops, but no controlled ablation isolates
this effect. Since F0 and F1 share the same template projection,
their performance gap cannot be attributed to that factor alone;
their search and validity/recovery policies also differ.
Projection-related domain mismatch therefore remains a possible,
unverified contributor to F1's lower performance.


\section{Implementation Details}
\label{app:implementation}

\paragraph{Dataset and Experimental Settings.}
All experiments follow the fixed PanoUAV-Track
training, validation, and test splits,
containing 1,160, 166, and 569 sequences,
respectively.
The test set contains 796 annotated target
tasks, of which 792 provide valid
post-initialization observations.
Each target is evaluated independently.
We consider two evaluation regimes.
In the \textbf{zero-shot} setting,
trackers are evaluated using their released
pretrained checkpoints without fine-tuning
on PanoUAV-Track.
In the \textbf{in-domain} setting,
trackers are trained using the PanoUAV-Track
training split with their respective
training objectives and architectures.

\begin{table*}[t]
\centering
\caption{
\textbf{In-domain training configurations on PanoUAV-Track.}
All models are trained for 100 epochs using AdamW.
LR denotes the base learning rate, and Batch/GPU
denotes the per-GPU training batch size.
}
\label{tab:training_config}

\vspace{-2mm}
\scriptsize
\setlength{\tabcolsep}{5pt}
\renewcommand{\arraystretch}{1.15}
\setlength{\arrayrulewidth}{0.4pt}

\resizebox{1.0\linewidth}{!}{%
\begin{tabular}{l|l|c|c|cc|l|c}
\hline
\rowcolor{gray!12}
Method &
Backbone Initialization &
Epochs &
Base LR &
Batch/GPU &
Global Batch &
LR Schedule &
Test Checkpoint \\
\hline

\rowcolor{gray!8}
\multicolumn{8}{l}{\textbf{Visual-only Trackers}} \\

OSTrack
& MAE ViT-B/16
& 100 & $4\times10^{-4}$
& 64 & 512
& $\times0.1$ at epoch 80
& 95 \\

ORTrack
& DeiT-T/16
& 100 & $4\times10^{-4}$
& 64 & 512
& $\times0.1$ at epoch 80
& 90 \\

SGLATrack
& Distilled DeiT-T/16
& 100 & $4\times10^{-4}$
& 64 & 512
& $\times0.1$ at epoch 80
& 100 \\

\hline
\rowcolor{gray!8}
\multicolumn{8}{l}{\textbf{Language-Initialized Trackers}} \\

JointNLT
& Swin-B + BERT
& 100 & $2\times10^{-3}$
& 8 & 64
& Warm-up; drops at 67/83/97
& 100 \\

MVLM
& Fast-iTPN-B + CLIP
& 100 & $5\times10^{-4}$
& 16 & 128
& Warm-up; drop at 90
& 100 \\

\hline
\rowcolor{gray!8}
\multicolumn{8}{l}{\textbf{Vision--Language Trackers}} \\

DUTrack
& Fast-iTPN-B/16
& 100 & $1\times10^{-4}$
& 16 & 128
& $\times0.1$ at epoch 89
& 100 \\

UVLTrack
& MAE ViT-B/16 + BERT
& 100 & $4\times10^{-4}$
& 16 & 128
& Cosine annealing
& 100\\

UETrack
& Fast-iTPN-T/16
& 100 & $1\times10^{-4}$
& 128 & 1024
& $\times0.1$ at epoch 80
& 100 \\

\hline
\end{tabular}%
}

\vspace{-2mm}
\end{table*}

\paragraph{Training Configuration.}
In-domain models use checkpoints selected
according to their respective training protocols,
as detailed in Table~\ref{tab:training_config}.
All models are trained for 100 epochs using AdamW,
while retaining their respective training objectives
and optimization schedules.
Generic pretrained visual and language encoders
are used where applicable, and UETrack additionally
employs a pretrained teacher for knowledge distillation.
Model-specific batch sizes and learning-rate schedules
are preserved to maintain compatibility with their
original implementations.

\paragraph{Initialization and Inference.}
Visual-only trackers use first-frame
bounding-box initialization.
Vision--language trackers additionally
receive a target referring expression.
Language-initialized trackers do not receive
a ground-truth initialization box but still
process video observations.

All baseline methods retain their native
image preprocessing, search-region construction,
and temporal update procedures.
For segmentation-based methods, predicted
masks are converted to bounding boxes
for consistent evaluation.

\paragraph{Geometry-Aware Adaptation.}
GASA is applied to the search-image branch
of existing tracking architectures.
LAPS adapts horizontal search coverage
according to target latitude, while SASC
performs periodic sampling across the
horizontal ERP seam.
The original template representation,
tracking network, and prediction head
remain unchanged.
GASA requires no additional optimization
and retains a single network forward pass
per tracking frame.

\section{Model-Specific Implementation Details}
\label{app:model_details}

We summarize the implementation configurations
of the evaluated tracking methods.
All models follow the common evaluation protocol
described in Sec.~\ref{app:implementation}.
For zero-shot evaluation, we use released
pretrained checkpoints without fine-tuning.
For in-domain evaluation, we preserve the
model-specific architectures and training
objectives while adapting them to PanoUAV-Track.

\subsection{Foundation Models}

\noindent\textbf{SAM 2.1}~\citep{ravi2025sam2}.
We use the Hiera-L variant with a
$1024\times1024$ input resolution.
The first-frame bounding box is supplied
as a visual prompt, followed by native
video mask propagation.
Predicted masks are converted into bounding
boxes for evaluation.

\noindent\textbf{SAM2Long}~\citep{ding2025sam2long}.
We adopt the Hiera-L backbone and retain
the original memory-tree inference strategy
with three propagation pathways.
The same initialization and mask-to-box
conversion protocol as SAM 2.1 is used.
No additional training is performed.

\noindent\textbf{SAM 3.1}~\citep{carion2026sam3}.
We evaluate the multiplex video predictor
using a $1008\times1008$ input resolution.
The initial bounding box is converted into
the native instance-prompt representation,
and subsequent masks are propagated
through the video predictor.
We use the SAM 3.1 Object Multiplex release
with the official $sam3.1_multiplex.pt$ checkpoint.

\subsection{Visual-Only Trackers}

\noindent\textbf{OSTrack}~\citep{ye2022joint}.
We adopt the ViT-B/16 configuration with
$128\times128$ template and $256\times256$
search inputs.
The corresponding crop factors are 2 and 4.
In-domain training initializes the backbone
from generic MAE pretraining.
The evaluation checkpoint is selected using
validation AUC.

\noindent\textbf{ORTrack}~\citep{wu2025ort}.
We use the DeiT-T/16 backbone with
$128\times128$ template and $256\times256$
search inputs.
The original occlusion-aware tracking
mechanism is retained.
In-domain training employs generic DeiT
initialization, with the evaluation checkpoint
selected on the validation split.

\noindent\textbf{SGLATrack}~\citep{xue2025sglatrack}.
We employ the distilled DeiT-T/16 backbone
with $128\times128$ template and
$256\times256$ search inputs.
The native similarity-guided layer-adaptive
mechanism is preserved.
In-domain training initializes the visual
encoder from pretrained DeiT weights.

\noindent\textbf{ODTrack}~\citep{zheng2024odtrack}.
We adopt the ViT-B/16 configuration with
$192\times192$ templates and a
$384\times384$ search image.
The original temporal token modeling
and memory-update mechanisms are retained.
Only zero-shot evaluation is reported.

\noindent\textbf{MCITrack}~\citep{kang2025exploring}.
We use the Fast-iTPN-B/16 backbone with
five $112\times112$ template representations
and a $224\times224$ search input.
The native contextual modeling and
temporal update procedures are preserved.
Only released pretrained weights are evaluated.

\noindent\textbf{MambaLCT}~\citep{li2025mambalct}.
We adopt the HiViT-B backbone with
two $128\times128$ templates and a
$256\times256$ search image.
The original long-term contextual modeling
and memory-update procedures are retained.
Evaluation uses the released checkpoint.

\noindent\textbf{LETrack}~\citep{letrack}.
We adopt the DeiT-Tiny/16 backbone with
$128\times128$ template and $256\times256$
search inputs.
The corresponding crop factors are 2 and 4.
The temporal representation contains 16 historical tokens.
The backbone is initialized from the distilled DeiT-Tiny
pretrained weights.
For evaluation, we directly use the official
\texttt{baseline\_WOCE} checkpoint from epoch 298,
without further fine-tuning on PanoUAV-Track.

\subsection{Language-Initialized Trackers}

\noindent\textbf{JointNLT}~\citep{zhou2023jointnlt}.
We adopt the Swin-B visual encoder and
BERT-based language encoder.
The initial target is identified through
full-image language grounding without
ground-truth box initialization.
Subsequent tracking follows the native
template-based inference procedure.
In-domain training uses pretrained visual
and language encoders.

\noindent\textbf{MVLM}~\citep{park2026mvlm}.
We evaluate the template-free variant
with a Fast-iTPN-B/16 visual backbone
and a CLIP-based language encoder.
The model performs language-guided
localization on full-image observations
resized to $224\times224$.
No ground-truth target box is supplied
during initialization.

\subsection{Vision--Language Trackers}

\noindent\textbf{DUTrack}~\citep{li2025dutrack}.
We employ the Fast-iTPN-B/16 backbone
with three $128\times128$ templates
and a $256\times256$ search image.
Both the target bounding box and referring
expression are provided.
The original dynamic language adaptation
and temporal tracking mechanisms are retained.
In-domain training initializes the visual
backbone from generic pretrained weights.

\noindent\textbf{UVLTrack}~\citep{ma2024uvltrack}.
We adopt a ViT-B/16 visual backbone and
BERT-based language encoder.
Template and search resolutions are
$128\times128$ and $256\times256$.
The native multimodal tracking and
template-update procedures are preserved.
In-domain training uses generic pretrained
visual and language encoders.

\noindent\textbf{UETrack}~\citep{kang2026uetrack}.
We employ the six-layer Fast-iTPN-T/16
visual encoder with $112\times112$
template and $224\times224$ search inputs.
Both visual and language prompts are provided.
In-domain training retains the original
teacher-based knowledge distillation protocol
using a pretrained tracking teacher.

\subsection{GASA Integration}

\noindent\textbf{MCITrack, ODTrack, ORTrack, and DUTrack with GASA}.
We integrate GASA into the search-image branch
of MCITrack, ODTrack, ORTrack, and DUTrack
while preserving their original tracking networks
and model-specific components.
The proposed adapter combines latitude-adaptive
horizontal search expansion with periodic
seam-aware sampling.
Predictions are mapped back into ERP coordinates
using geometry-consistent coordinate recovery.
For zero-shot evaluation, GASA is directly
applied to the released pretrained models
of MCITrack, ODTrack, ORTrack, and DUTrack.
For in-domain evaluation, GASA is applied
to the corresponding PanoUAV-Track-trained
ORTrack and DUTrack models without additional
optimization.
Each baseline--GASA pair uses the same checkpoint.
Both configurations preserve a single
tracking-network forward pass per frame.

\noindent\textbf{Effectiveness across trackers.}
Table~\ref{tab:gasa_geometry_comparison} shows
consistent improvements across all four trackers.
Under zero-shot evaluation, GASA improves the
overall AUC of MCITrack, ODTrack, ORTrack, and
DUTrack by 9.04, 9.10, 8.74, and 9.36 percentage
points, respectively.
The gains also persist under in-domain evaluation:
GASA improves ORTrack from 74.06 to 82.31 AUC
and DUTrack from 76.61 to 82.84 AUC.
Since each baseline--GASA pair uses the same
checkpoint, these improvements demonstrate
the benefit of the inference-time adapter
without additional model optimization.

\noindent\textbf{Seam robustness.}
The largest improvements occur within seam-event
windows.
In the zero-shot setting, GASA increases Seam AUC
by 48.84, 47.25, 46.07, and 47.44 percentage points
for MCITrack, ODTrack, ORTrack, and DUTrack,
respectively.
Meanwhile, the absolute gap between Seam and
Non-Seam AUC decreases from 38.03--41.45
percentage points to at most 1.48 percentage points.
A similar trend holds for the in-domain models:
Seam AUC improves by 43.59 percentage points
for ORTrack and 42.42 percentage points for DUTrack,
while their seam deficits decrease from 38.99
and 41.83 to 1.40 and 3.18 percentage points,
respectively.
These results are consistent with improved
robustness to ERP seam discontinuities.

\noindent\textbf{Performance across latitudes.}
GASA improves AUC in all three latitude groups
for every evaluated tracker in both settings.
For the zero-shot models, the gains range from
9.80 to 11.61 percentage points in Equator,
6.68 to 7.30 in Middle, and 7.04 to 8.57 in Polar.
The Middle--Equator gap also decreases for all
evaluated models in both settings.
However, the gains are not confined to high
latitudes, and performance differences across
latitude groups remain.
The results therefore support broad improvements
across latitude conditions.


\section{Step-by-Step Illustration of GASA}
\label{app:gasa_example}

\begin{figure*}[t!]
    \centering
    \includegraphics[width=\linewidth]{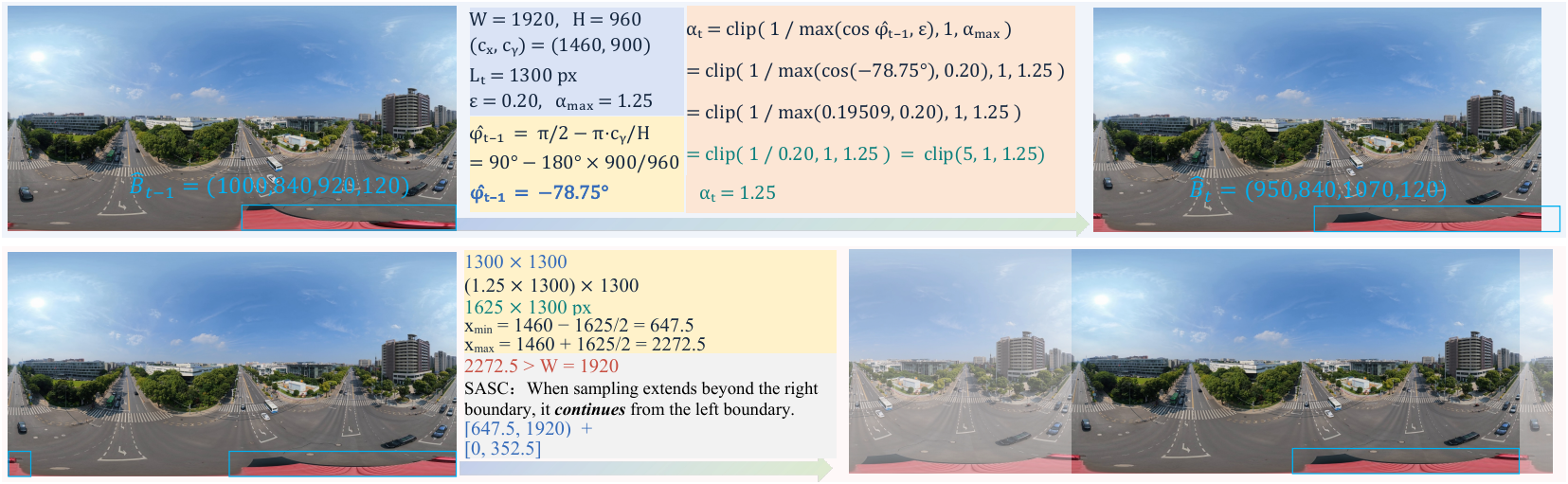}
    \vspace{-5.5mm}
    \caption{
    \textbf{Step-by-step illustration of GASA.}
    Given the previous target prediction,
    LAPS determines the latitude-dependent
    horizontal search expansion.
    SASC constructs a continuous search region
    by periodically sampling pixels across
    the horizontal ERP boundary.
    The adapted search image is processed by
    the unchanged tracking network, followed
    by geometry-consistent coordinate recovery
    to preserve seam-crossing bounding boxes.
    }
    \label{fig:gasa_example}
    \vspace{-1.6em}
\end{figure*}

To provide an intuitive understanding of GASA,
we present a numerical example illustrating
latitude-adaptive search expansion, seam-aware
periodic sampling, and geometry-consistent
bounding-box recovery.
As shown in \cref{fig:gasa_example},
the example considers a horizontally elongated
target located near the bottom of an ERP image.
The numerical values are illustrative and
demonstrate the complete processing pipeline.

\paragraph{Step 1: Search Region Initialization.}
Consider an ERP frame with resolution
$W\times H=1920\times960$.
The previous predicted bounding box is

\begin{equation}
\widehat B_{t-1}
=
(1000,840,920,120),
\end{equation}

where the four values denote the top-left
corner and bounding-box dimensions.

The corresponding target center is

\begin{equation}
\begin{aligned}
c_x &=1000+\frac{920}{2}=1460,\\
c_y &=840+\frac{120}{2}=900.
\end{aligned}
\end{equation}

Following the original tracker, we consider
a square search region with side length
$L_t=1300$ pixels.

The predicted target latitude is

\begin{equation}
\begin{aligned}
\widehat{\phi}_{t-1}
&=\frac{\pi}{2}-\pi\frac{c_y}{H}\\
&=90^\circ-180^\circ\times\frac{900}{960}\\
&=-78.75^\circ.
\end{aligned}
\end{equation}

The target is therefore located near the
southern polar region of the ERP image,
where horizontal projection distortion
is relatively pronounced.

\paragraph{Step 2: Latitude-Adaptive Expansion.}
Given $\epsilon=0.20$ and
$\alpha_{\max}=1.25$, LAPS computes the
horizontal expansion factor as

\begin{equation}
\begin{aligned}
\alpha_t
&=
\operatorname{clip}
\left(
\frac{1}{
\max(\cos\widehat{\phi}_{t-1},\epsilon)
},
1,\alpha_{\max}
\right)\\
&=
\operatorname{clip}
\left(
\frac{1}{\max(0.19509,0.20)},
1,1.25
\right)\\
&=\operatorname{clip}(5,1,1.25)\\
&=1.25.
\end{aligned}
\end{equation}

Accordingly, the horizontal search extent
increases from $1300$ to

\begin{equation}
\alpha_tL_t
=
1.25\times1300
=
1625.
\end{equation}

The vertical search extent remains
unchanged at $1300$ pixels.

Using Eq.~\eqref{eq:gasa_sampling},
the adapted search boundaries are

\begin{equation}
\begin{aligned}
x_{\min}
&=1460-\frac{1625}{2}=647.5,\\
x_{\max}
&=1460+\frac{1625}{2}=2272.5,\\
y_{\min}
&=900-\frac{1300}{2}=250,\\
y_{\max}
&=900+\frac{1300}{2}=1550.
\end{aligned}
\end{equation}

Thus, LAPS constructs a search region
with horizontal and vertical extents
of $1625$ and $1300$ pixels, respectively.
The search center remains unchanged,
and the final network input resolution
is preserved.
Vertical coordinates outside the image
are handled using the original tracker's
padding policy.


\begin{table}[t]
\centering
\providecolor{darkred}{RGB}{160,30,30}
\providecolor{darkgreen}{RGB}{0,100,50}

\newcommand{\GASApos}[1]{\textcolor{darkred}{#1}}
\newcommand{\GASAneg}[1]{\textcolor{darkgreen}{#1}}
\newcommand{\GASAcell}[2]{%
  \shortstack{#1\\{\tiny(#2)}}%
}

\caption{
\textbf{Geometry-aware GASA comparison on PanoUAV-Track.}
All scores are AUC (\%). Each pair uses the same checkpoint.
Zero-shot models use official pretrained weights without fine-tuning;
in-domain models use PanoUAV-Track-trained weights.
Equator, Middle, and Polar correspond to absolute GT-center latitudes
$<30^\circ$, $[30^\circ,60^\circ)$, and $\geq60^\circ$.
Seam denotes $\pm15$-frame windows around consecutive visible
GT-center jumps $|c_{x,t}-c_{x,t-1}|>W/2$.
}

\label{tab:gasa_geometry_comparison}
\vspace{-2mm}
\scriptsize
\setlength{\tabcolsep}{3pt}
\renewcommand{\arraystretch}{1.05}
\setlength{\arrayrulewidth}{0.4pt}

\resizebox{\columnwidth}{!}{%
\begin{tabular}{l|c|ccccc|ccc}

\noalign{\global\setlength{\arrayrulewidth}{0.6pt}}
\hline
\noalign{\global\setlength{\arrayrulewidth}{0.4pt}}

\rowcolor{gray!10}
& & \multicolumn{5}{c|}{Latitude Distortion}
& \multicolumn{3}{c}{Seam Discontinuity} \\

\rowcolor{gray!10}
Method & Overall & Equator & Middle & Polar
& M$-$E & P$-$M
& Non-Seam & Seam & Drop \\

\hline
\rowcolor{gray!8}
\multicolumn{10}{l}{
\textbf{Zero-shot: official pretrained weights}
} \\

MCITrack~\citep{kang2025exploring}
& 77.63
& 75.79
& 80.50
& 76.61
& \GASApos{+4.71}
& \GASAneg{-3.89}
& 79.31
& 37.86
& \GASAneg{-41.45} \\

MCITrack+\textcolor{darkred}{\textbf{GASA}}
& \GASAcell{86.67}{\GASApos{+9.04}}
& \GASAcell{85.59}{\GASApos{+9.80}}
& \GASAcell{87.18}{\GASApos{+6.68}}
& \GASAcell{83.65}{\GASApos{+7.04}}
& \GASAcell{\GASApos{+1.59}}{\GASAneg{-3.12}}
& \GASAcell{\GASAneg{-3.53}}{\GASApos{+0.36}}
& \GASAcell{86.09}{\GASApos{+6.78}}
& \GASAcell{86.70}{\GASApos{+48.84}}
& \GASAcell{\GASApos{+0.61}}{\GASApos{+42.06}} \\

\hline

ODTrack~\citep{zheng2024odtrack}
& 76.55
& 71.81
& 78.77
& 77.94
& \GASApos{+6.96}
& \GASAneg{-0.83}
& 76.90
& 35.82
& \GASAneg{-41.08} \\

ODTrack+\textcolor{darkred}{\textbf{GASA}}
& \GASAcell{85.65}{\GASApos{+9.10}}
& \GASAcell{82.61}{\GASApos{+10.80}}
& \GASAcell{86.07}{\GASApos{+7.30}}
& \GASAcell{85.66}{\GASApos{+7.72}}
& \GASAcell{\GASApos{+3.46}}{\GASAneg{-3.50}}
& \GASAcell{\GASAneg{-0.41}}{\GASApos{+0.42}}
& \GASAcell{84.55}{\GASApos{+7.65}}
& \GASAcell{83.07}{\GASApos{+47.25}}
& \GASAcell{\GASAneg{-1.48}}{\GASApos{+39.60}} \\

\hline

ORTrack~\citep{wu2025ort}
& 72.74
& 67.41
& 76.44
& 66.96
& \GASApos{+9.03}
& \GASAneg{-9.48}
& 72.66
& 34.63
& \GASAneg{-38.03} \\

ORTrack+\textcolor{darkred}{\textbf{GASA}}
& \GASAcell{81.48}{\GASApos{+8.74}}
& \GASAcell{79.02}{\GASApos{+11.61}}
& \GASAcell{83.38}{\GASApos{+6.94}}
& \GASAcell{75.53}{\GASApos{+8.57}}
& \GASAcell{\GASApos{+4.36}}{\GASAneg{-4.67}}
& \GASAcell{\GASAneg{-7.85}}{\GASApos{+1.63}}
& \GASAcell{80.64}{\GASApos{+7.98}}
& \GASAcell{80.70}{\GASApos{+46.07}}
& \GASAcell{\GASApos{+0.06}}{\GASApos{+38.09}} \\

\hline

DUTrack~\citep{li2025dutrack}
& 75.75
& 71.15
& 78.12
& 77.76
& \GASApos{+6.97}
& \GASAneg{-0.36}
& 76.26
& 36.00
& \GASAneg{-40.26} \\

DUTrack+\textcolor{darkred}{\textbf{GASA}}$^{\dagger}$
& \GASAcell{85.11}{\GASApos{+9.36}}
& \GASAcell{82.49}{\GASApos{+11.34}}
& \GASAcell{85.18}{\GASApos{+7.06}}
& \GASAcell{85.07}{\GASApos{+7.31}}
& \GASAcell{\GASApos{+2.69}}{\GASAneg{-4.28}}
& \GASAcell{\GASAneg{-0.11}}{\GASApos{+0.25}}
& \GASAcell{84.00}{\GASApos{+7.74}}
& \GASAcell{83.44}{\GASApos{+47.44}}
& \GASAcell{\GASAneg{-0.56}}{\GASApos{+39.70}} \\

\hline
\hline

\rowcolor{gray!8}
\multicolumn{10}{l}{
\textbf{In-domain: PanoUAV-Track-trained weights}
} \\

ORTrack~\citep{wu2025ort}
& 74.06
& 61.90
& 77.79
& 77.35
& \GASApos{+15.89}
& \GASAneg{-0.44}
& 71.96
& 32.97
& \GASAneg{-38.99} \\

ORTrack+\textcolor{darkred}{\textbf{GASA}}
& \GASAcell{82.31}{\GASApos{+8.25}}
& \GASAcell{70.69}{\GASApos{+8.79}}
& \GASAcell{83.61}{\GASApos{+5.82}}
& \GASAcell{83.57}{\GASApos{+6.22}}
& \GASAcell{\GASApos{+12.92}}{\GASAneg{-2.97}}
& \GASAcell{\GASAneg{-0.04}}{\GASApos{+0.40}}
& \GASAcell{77.96}{\GASApos{+6.00}}
& \GASAcell{76.56}{\GASApos{+43.59}}
& \GASAcell{\GASAneg{-1.40}}{\GASApos{+37.59}} \\

\hline

DUTrack~\citep{li2025dutrack}
& 76.61
& 70.92
& 80.17
& 81.62
& \GASApos{+9.25}
& \GASApos{+1.45}
& 77.55
& 35.72
& \GASAneg{-41.83} \\

DUTrack+\textcolor{darkred}{\textbf{GASA}}
& \GASAcell{82.84}{\GASApos{+6.23}}
& \GASAcell{77.06}{\GASApos{+6.14}}
& \GASAcell{84.21}{\GASApos{+4.04}}
& \GASAcell{85.75}{\GASApos{+4.13}}
& \GASAcell{\GASApos{+7.15}}{\GASAneg{-2.10}}
& \GASAcell{\GASApos{+1.54}}{\GASApos{+0.09}}
& \GASAcell{81.32}{\GASApos{+3.77}}
& \GASAcell{78.14}{\GASApos{+42.42}}
& \GASAcell{\GASAneg{-3.18}}{\GASApos{+38.65}} \\

\noalign{\global\setlength{\arrayrulewidth}{0.6pt}}
\hline
\noalign{\global\setlength{\arrayrulewidth}{0.4pt}}

\end{tabular}%
}

\vspace{0.5mm}
\parbox{\columnwidth}{\scriptsize\textit{Notes.}
Overall averages targets equally; geometry columns pool valid
target-frames, excluding initialization and invisible GT.
M$-$E = Middle $-$ Equator;
P$-$M = Polar $-$ Middle;
Drop = Seam $-$ Non-Seam.
Parentheses show GASA minus its baseline in percentage points.
Signed differences are computed from displayed two-decimal values;
positive values are red and negative values are green.
GASA uses latitude-adaptive LAPS
($\epsilon=0.20$, $\alpha_{\max}=1.25$)
and periodic bilinear SASC.
$^{\dagger}$The reused zero-shot DUTrack result uses
$\alpha_{\max}=1.5$.
The event-window Seam score is distinct from
seam-frame $\mathrm{AUC}_{\mathrm{S}}$.
}

\end{table}

\paragraph{Step 3: Seam-Aware Periodic Sampling.}
Since the horizontal search boundary
satisfies

\begin{equation}
x_{\max}=2272.5>W=1920,
\end{equation}

the adapted search region extends beyond
the right ERP boundary.

Unlike conventional planar padding,
SASC exploits the horizontal periodicity
of ERP by mapping the sampling coordinates
according to

\begin{equation}
x_{\mathrm{wrap}}=x\bmod W.
\end{equation}

Consequently, the original horizontal
sampling interval $[647.5,2272.5]$
is reconstructed from two source regions:

\begin{equation}
\underbrace{[647.5,1920)}_{\text{right ERP region}}
\;\cup\;
\underbrace{[0,352.5]}_{\text{left ERP region}}.
\end{equation}

The two intervals are spatially adjacent
on the viewing sphere, despite appearing
at opposite sides of the ERP image.

Using periodic bilinear interpolation,
SASC combines these regions into a
single continuous search image.
Fractional sampling coordinates near
the ERP boundary are interpolated
using neighboring pixels from both
sides of the image.

The adapted region is resampled to
the original network input resolution
and processed using a single forward
pass of the pretrained tracker.

\paragraph{Step 4: Geometry-Consistent Recovery.}
Suppose the tracking head predicts the
following normalized bounding box
in search-image coordinates:

\begin{equation}
\widehat B_t^s
\approx
(0.186154,0.453846,0.658462,0.092308),
\end{equation}

where the values represent the normalized
top-left corner and box dimensions.

These values correspond to the adapted
search region determined by LAPS.
Using Eq.~\eqref{eq:gasa_recovery},
the prediction is mapped back to ERP
coordinates as follows:

\begin{equation}
\begin{aligned}
\widehat{x}_t
&=
\left[
1460+1625(0.186154-0.5)
\right]\bmod1920 \\
&\approx950.
\end{aligned}
\end{equation}

\begin{equation}
\begin{aligned}
\widehat{y}_t
&=
900+1300(0.453846-0.5) \\
&\approx840.
\end{aligned}
\end{equation}

\begin{equation}
\begin{aligned}
\widehat{w}_t
&=
1625\times0.658462 \\
&\approx1070.
\end{aligned}
\end{equation}

\begin{equation}
\begin{aligned}
\widehat{h}_t
&=
1300\times0.092308 \\
&\approx120.
\end{aligned}
\end{equation}

Therefore, the recovered ERP prediction is

\begin{equation}
\boxed{
\widehat B_t=(950,840,1070,120).
}
\end{equation}

This example demonstrates that the
inverse search transformation correctly
recovers both the target position and
its bounding-box dimensions.

\paragraph{Step 5: Seam-Crossing Box Representation.}
The recovered bounding box satisfies

\begin{equation}
\widehat{x}_t+\widehat{w}_t
=
950+1070
=
2020>1920.
\end{equation}

Hence, the predicted target extends
across the horizontal ERP seam.

For visualization, its visible support
is divided into two regions:

\begin{equation}
\begin{aligned}
B_{\mathrm{right}}
&=(950,840,970,120),\\
B_{\mathrm{left}}
&=(0,840,100,120).
\end{aligned}
\end{equation}

The right region contains $970$ pixels
of horizontal target extent, while
the remaining $100$ pixels appear
at the left ERP boundary.

These two visible regions represent
a single continuous target.
Importantly, the recovered bounding-box
width remains $1070$ pixels rather
than being truncated at the image boundary.

The predicted target center is

\begin{equation}
\begin{aligned}
\widehat c_x
&=
\left(
950+\frac{1070}{2}
\right)\bmod1920
=1485,\\
\widehat c_y
&=
840+\frac{120}{2}
=900.
\end{aligned}
\end{equation}

The recovered prediction is subsequently
used to construct the search region
for the next frame, without requiring
additional ground-truth annotations.

The example illustrates the complementary
roles of LAPS and SASC.
LAPS expands horizontal search coverage
according to the predicted target latitude,
while SASC preserves spatial continuity
when the search region crosses the ERP seam.
Geometry-consistent recovery then maps
the predicted bounding box back to ERP
coordinates without losing its
cross-boundary representation.
All operations are performed at the
search-region construction and
coordinate-recovery stages, preserving
the original tracking architecture
and requiring no additional network
forward passes.


\end{document}